\documentclass{article}
\usepackage{nips15submit_e}

\usepackage{times}
\usepackage{amssymb}
\usepackage{amsmath}
\usepackage{amsthm}
\usepackage{bm}
\usepackage{tikz}
\usepackage{tkz-euclide}
\usetkzobj{all}
\usepackage{chngcntr}
\usepackage{verbatim}
\usepackage{mathtools}
\usepackage{esvect}
\RequirePackage{algorithm}
\RequirePackage{algorithmic}
\usepackage{thmtools}

\newcommand{\x}{{\bm{x}}} 
\newcommand{\z}{{\bm{z}}} 

\newcommand{\E}{{\mathbf{E}}} 
\newcommand{\bound}{{\cal H}} 
\renewcommand{\S}{{\cal S}} 
\newcommand{\T}{{\cal T}} 
\renewcommand{\Re}{{\mathbb R}} 
\newcommand{\fclass}{{\mathcal F}} 

\DeclareMathOperator*{\argmin}{\arg\min}

\newtheorem{theorem}{Theorem}
\newtheorem{lemma}[theorem]{Lemma}

\newtheorem{corollary}[theorem]{Corollary}

\declaretheoremstyle[notefont=\bfseries,notebraces={}{},%
    headpunct={},postheadspace=0.4em]{mystyle}
\declaretheorem[style=mystyle,numbered=no,name=Lemma]{LemmaM}
\declaretheorem[style=mystyle,numbered=no,name=Theorem]{TheoremM}
\declaretheorem[style=mystyle,numbered=no,name=Corollary]{CorollaryM}

\graphicspath{ {./images/} }

\newcommand{\ignore}[1]{}

\author{Hadi Daneshmand, Aurelien Lucchi, Thomas Hofmann \\
 Department of Computer Science, ETH Zurich, Switzerland
 }

\newcolumntype{S}{>{\centering\arraybackslash} m{.10\linewidth} }
\newcolumntype{T}{>{\centering\arraybackslash} m{.30\linewidth} }

\title{ { {\bf \sc DynaNewton}}\\ Accelerating Newton's Method for Machine Learning}

\nipsfinalcopy

\begin{document}
\maketitle

\begin{abstract}
Newton's method is a fundamental technique in optimization with  quadratic convergence within a neighborhood around the optimum. However reaching this neighborhood is often slow and dominates the computational costs. We exploit two properties specific to empirical risk minimization problems to accelerate Newton's method, namely, subsampling training data and increasing strong convexity through regularization. We propose a novel continuation method, where we define a family of objectives over increasing sample sizes and with decreasing regularization strength. Solutions on this path are tracked  such that the minimizer of the previous objective is guaranteed to be within the  quadratic convergence region of the next objective to be optimized. Thereby every Newton iteration is guaranteed to achieve super-linear contractions with regard to the chosen objective, which becomes a moving target.
We provide a theoretical analysis that motivates our algorithm, called {\sc DynaNewton}, and characterizes its speed of convergence. Experiments on a wide range of data sets and problems consistently confirm the predicted computational savings. 
\end{abstract}

\section{Introduction}

In machine learning, we often fix a function class with parameters $\x \in \Re^d$, define a non-negative family of loss functions $\phi^z$, a regularizer $\Omega$, and then aim to minimize a regularized sample loss over training data $\S$,
\begin{align}\label{eq:ERM}
& f^\S_{\nu}(\x) :=  \frac {1}{|\S|} \sum_{\z \in \S} \phi_\nu^\z(\x), \quad \phi_\nu^\z(\x) := \phi^\z(\x) +   \nu \Omega(\x) \,.
\end{align}
Justified by theories like Tikhonov regularization or structural risk minimization, we know that we can control the expected risk of the minimizers $\x^*_\nu$ of $f^\S_{\nu}$, i.e.~avoid overfitting, by choosing the regularization strength $\nu$ appropriately.

In this paper, we focus on Newton's method for optimization, which obeys quadratic convergence towards an extremal point, when initialized sufficiently close to such a point \cite{ortega1970iterative}. However, despite this unmatched speed of convergence, Newton's method has shortcomings for large-scale machine learning problems of the type described above: (i) Reaching the quadratic convergent regime through an initial \textit{damped phase} \cite{boyd2004convex} is where most of the computation is typically spent, unless one has access to an initial solution close-enough to the optimum. (ii) Being a batch algorithm, each Newton iteration involves a complete pass over the entire data set to compute the required gradient and Hessian matrix. (iii) The computation of the Newton update requires to solve a linear system of equations, which is a challenge, in particular, if the data dimensionality is high. 

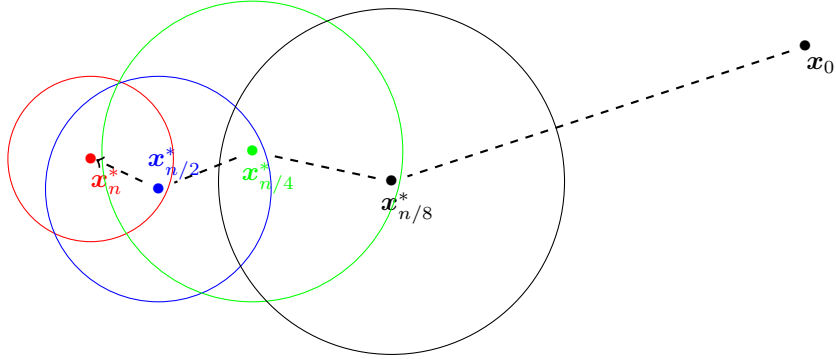
\begin{figure}[t]
\vspace*{5mm}
    \begin{center}
        \begin{tikzpicture}
        \node (A) at (-0.5,1.5) {\color{red}\textbullet};
        \tkzLabelSegment[below=0.2pt](A,A){\color{red}\textit{$\x^*_{n}$}}
          \node (F) at (-0.55,1.55) {};
         \node (B) at (0.4,1.1) {\color{blue}\textbullet};
         \tkzLabelSegment[above=0.2pt](B,B){\textit{$\color{blue}\x^*_{n/2}$}}
	 \node (C) at (1.65,1.6) {\color{green}\textbullet};
         \tkzLabelSegment[below=0.5pt](C,C){\textit{$\color{green}\x^*_{n/4}$}}
          \node (D) at (3.5,1.2) {\textbullet};
         \tkzLabelSegment[below=0.5pt](D,D){\textit{$\x^*_{n/8}$}}
           \node (E) at (9.0,3.0) {\color{black}\textbullet};
         \tkzLabelSegment[below=0.5pt](E,E){\textit{$\color{black}\x_0$}}
        \draw  [red] (A) circle (1.1cm);
        \draw  [blue] (B) circle (1.5cm);
         \draw  [green] (C) circle (2.0cm);
         \draw  [black] (D) circle (2.3cm);
         \draw  [->,dashed,thick] (E) -- (D) -- (C) -- (B) -- (F);
    \end{tikzpicture}
    \end{center}
\label{figure:sketch}
\caption{Illustration of the continuation method approach proposed in this paper: Assume we chose $\nu \propto 1/n$. After optimizing the risk for a small subsample, we continuously increase the sample size (by a factor, e.g.~$2$) and proportionally lower $\nu$. We want to guarantee that each solution provides a starting point that is within the quadratic convergence region of the subsequent optimization. These regions are depicted by colored circles.}
\end{figure}

The strategy presented in this paper is as follows. Following the ideas of \cite{daneshmand2016small}, we present a systematic way to dynamically subsample the data  so as to match-up statistical and computational accuracy, leading to significant savings in the amount of overall computation. Moreover, employing this together with an adaptive control of the regularization strength, we define a \textit{continuation method} \cite{allgower2012numerical}, where we track solutions computed for problems with fewer data and with stronger regularization. We use previous solutions in order to compute the starting point for the next Newton iteration(s), possibly operating on a larger sample. An ideal sketch of the situation is shown in Figure \ref{figure:sketch}. This is meant to address challenges (i) and (ii). Finally, in order to overcome challenge (iii), we also empirically investigate our strategy for a popular quasi-Newton method, namely BFGS, which computes approximations to the inverse Hessian through closed-form rank-one matrix updates.

\section{Related work}

\paragraph{Adaptive data sub-sampling}

Recent results have shown that the finite sum structure of the empirical risk can be exploited to achieve linear convergence for strongly convex objectives~\cite{johnson2013accelerating, defazio2014saga, wang2016reducing}. 
This is accomplished by revisiting samples and storing their gradients in order to reduce the variance of future update directions. Although these methods achieve fast convergence on the empirical risk, they do not explicitly consider the expected risk, which has been separately studied in the literature on learning theory. It is usually analysed with the help of uniform convergence bounds that take the generic form  \cite{boucheron2005theory}
\begin{align}
\E_\S \left[ \sup_{\x \in {\cal F}} \left| f^\S(\x) - \E_\z \phi^\z(\x) \right| \right]  \leq
\bound(n)\,,
\label{eq:bound}
\end{align} 
where the expectation is over a random sample $\S$ of size $n$. Here $\bound$ is a bound that depends on $n$, usually through a ratio $n/d$, where $d$ is the capacity of $\fclass$ (e.g.~VC dimension)~\cite{bousquet2008tradeoffs}.\footnote{For ease of exposition, we assume that all regularized risks $f_\nu$ can be governed by the same function $\bound$.}

The recent work of~\cite{daneshmand2016small} simultaneously exploits the properties of the empirical risk as well as the concentration bounds from learning theory to achieve fast convergence to the expected risk. This approach uses a dynamic sample size schedule that  matches up optimization error and statistical accuracy. Although this approach was tailored specifically for variance-reduced stochastic gradient methods, we here show how a similar adaptive sample size strategy can be used in the context of non-stochastic approaches such as Newton's method.

\paragraph{Regularization paths and continuation methods}

There is a rich body of literature on numerical continuation methods, a reference work being~\cite{allgower2012numerical}. The basic idea is to define a family of objectives, in the simplest case with a single parameter $t \in[0;T]$, and to optimize over a sequence of objectives $f_t$ with increasing $t$, such that the sequence approaches some desired final $f= f_T$. The general motivation is that following the solution path $\x^*_t = \argmin_\x f_t(\x)$ may be computationally more efficient than optimizing the (typically) harder problem $f$ directly. 

Free energy based continuation methods, often for non-convex or integer problems, have been popularized in computer vision and machine learning under the name of \textit{deterministic annealing}~\cite{rose1998deterministic}, a deterministic variant of  simulated annealing~\cite{kirkpatrick1984optimization}. Here the family of objectives is parametrized by the computational analogue of temperature. Similar techniques known as graduated optimization have also been proposed in computer vision~\cite{blake1987visual} and in machine learning, a recent example being~\cite{hazan2015graduated}. 

In machine learning, the model complexity is typically controlled through a regularizer. In structural risk minimization, choosing a good regularizer plays an important role in the bias-variance trade-off for model selection \cite{vapnik1998statistical}. Another virtue of the regularization factor is its influence on the performance of the optimization procedure, as can be seen through continuation methods or \textit{regularization path} techniques~\cite{hastie2004entire, bach2005computing}. This is formally described by defining a family of objectives functions $f_t := f_{\nu_t}$ with a decreasing sequence $\nu_t$ which progressively provide a better approximation of $f$. The typical goal pursued in this line of work is to combine optimization with model selection, although \cite[Section 4.3]{hastie2004entire} also report computational savings. Our use of a continuation method is purely motivated by computational complexity and justified by a rigorous analysis of the quadratic convergence regime of Newton's method as described in detail in the next section.

\section{Adaptive Newton Method}

\subsection{Newton's Method}

Assume that we have a $\mu$ strongly-convex function $f: \Re^d \to \Re$, which we want to minimize over solutions $\x \in \Re^d$. A Newton step defines the increment as 
\begin{align}
\triangle \x = - \left[ \nabla^2 f(\x) \right]^{-1} \nabla f(\x), \quad \x \leftarrow \x + \triangle \x
\label{eq:invhessian}
\end{align}
An equivalent way to define Newton increments without the need to invert the Hessian is implicitly as the solution of the linearized optimality condition 
\begin{align}
\nabla f( \x + \triangle\x) \approx \nabla f(\x) + \nabla^2 f(\x) \triangle \x \stackrel != 0
\end{align}
as can be verified by plugging in Eq.~\eqref{eq:invhessian}. 

Newton's method converges to the optimal solution $\x^* := \arg \min f(\x)$ in a finite number of steps. The speed of convergence is characterized by two distinct phases that depend on the distance to $\x^*$. The first phase is a {\it damped phase} with slow convergence while the second phase has quadratic convergence and is triggered when entering a region close to $\x^*$.

In order to formally characterized this region of quadratic convergence, an important quantity is the \textit{Newton decrement} function~\cite{boyd2004convex} defined as
\begin{align}
\lambda_f := \sqrt{\nabla f^\top \left[ \nabla^2 f \right]^{-1} \nabla f}, \quad \lambda_f: \Re^d \to \Re_{\ge 0}\,.
\end{align}
We will directly make use of this definition in conjunction with an additional requirement that we impose on $f$, namely that it is \textit{self-concordant} \cite{nesterov1994interior}. Self-concordance allows for an elegant, affine-invariant characterization of the quadratic convergence region, as detailed in \cite{boyd2004convex}, leading to the sufficient condition that $\lambda_f(\x) \le \eta$, where $\eta \in(0; \tfrac 14)$ is a constant whose exact value depends on control parameters for the line search. The self-concordance property restricts the set of functions $f$, yet it is known that in practice, a similar analysis often optimistically applies to a wider range of functions. For further details, we refer the reader to the discussion in \cite[Section 9.6]{boyd2004convex} and the work on logistic regression in~\cite{bach2010self}. 

Note that strong convexity implies $\nabla^2 f \succeq \mu \mathbf I$ and thus $[\nabla^2 f]^{-1} \preceq \frac 1 \mu \mathbf I$, so that immediately
\begin{align}
\lambda_f(\x) 
\le \| \nabla f(\x)\|_{\frac 1 \mu \mathbf I} = \frac 1 {\sqrt \mu} \| \nabla f(\x)\| \;\!\Longrightarrow\!\; \| \nabla f(\x)\| \stackrel !\le \eta  \sqrt{\mu} 
\label{eq:norm-condition} 
\end{align} 
and we arrive at a simple (sufficient) condition on the gradient norm at $\x$.

\subsection{Continuation Method} 

We study the $2$-parametric family of regularized empirical risk functions that are defined via smooth and convex, non-negative loss functions $\phi^\z$ and relative to a full sample $\S^*=(\z_1,\dots,\z_N)$. We make use of the definitions in Eq.~\eqref{eq:ERM} and we think of $f^\S_\nu$ as being indexed by $\nu$ as well as $n = |\S|$, where $\S = (\z_1,\dots,\z_n)$ consists of the first $n$ samples of $\S^*$. The canonical regularizer we consider is $\Omega(\x) = \frac 12 \| \x \|^2$ and we will sometimes utilize the commonly used heuristics of choosing $\mu \propto 1/m$ to focus on a simpler $1$-parametric family. 

We want to implement the abstract procedure described in Algorithm \ref{alg:continuation}, where we either pre-generate a sequence of problems $f_t$ or, alternatively, construct $(\mu_t,m_t)$ greedily in a data-adaptive manner.
\begin{algorithm}[t]
\begin{algorithmic}[1]
\STATE given sample $\S$, iterations $T$ 
\STATE {\bf V1}: given sequence $(\mu_t, m_t)$, 
 {\bf V2}: given starting point $(\mu_0, m_0)$
\STATE $\x_0 \leftarrow \argmin_\x f_{\mu_0,m_0}(\x;\S)$ 
\FOR{$t=1,\dots,T$}
\STATE {\bf V1}: do nothing, 
{\bf V2}: compute $\mu_{t}$ and $m_{t}$
\STATE compute Newton increment $\triangle \x$ for $f_t(\x) := f_{\mu_t,m_t}(\x_{t-1},\S)$
\STATE $\x_t \leftarrow \x_{t-1} + \triangle \x$
\ENDFOR
\end{algorithmic}
\caption{\label{alg:continuation} Basic Newton continuation method: \textit{a priori} (V1) and \textit{data-adaptive} variant (V2).} 
\end{algorithm}
The key condition for making this  work as expected (by design) is that we are able to establish the following condition 
\begin{align}
\label{eq:condition-handover}
\lambda_{f_t}(\x^*_{t-1}) \le \eta \quad \text{or, more conservatively,} \quad \| \nabla f_{t}(\x^*_{t-1})\| \le \eta \sqrt{\mu_t} \,,
\end{align}
which will assure (under appropriate assumptions, e.g.~self-concordance) that the minimizer of the previous optimization problem will provide a starting point that is within the quadratic convergence region of the subsequent optimization problem, yielding a proper "hand-over" of the solution as illustrated in Figure \ref{figure:sketch}. 

\subsection{Reducing Regularization Strength}

Let us first assume that we fix $\S$ and that $\mu:= \mu_{t-1}$. We are seeking for the range of possible $\nu := \mu_{t} \le \mu$ for which we can guarantee the condition in Eq.~\eqref{eq:condition-handover}. 

\begin{lemma}
\label{lemma:erm-norm}
Let $\Omega(\cdot) = \frac 12 \| \cdot \|^2$, $\x^*_\mu := \arg\min_\x f_{\mu}(\x)$, and $B_\mu := \frac{\eta^2}{\mu \| \x^*_\mu\|^2}$. For any $\nu$ such that 
\begin{align}
\mu \ge \nu \ge \mu \left( 1 -  \frac 12 \left( \sqrt{B_\mu^2 + 4B_\mu} - B_\mu\right) \right) \Longrightarrow 
\lambda_{f_\nu}(\x^*_\mu)  \le  \eta
\label{eq:nu-bound}
\end{align}
\begin{proof}
By the first order optimality condition for $\x^*_\mu$, we have that $\nabla f_0(\x^*_\mu) = - \mu \x^*_\mu$, hence
\begin{align}
\nabla f_\nu(\x^*_\mu) = 
\nabla f_0(\x^*_\mu) + \nu \x^*_\mu = 
(\nu - \mu) \x^*_\mu \quad \Longleftrightarrow \quad
 \| \nabla f_\nu(\x^*_\mu) \| = (\mu-\nu) \| \x^*_\mu\| \,.
\label{eq:simplenormeq}
\end{align}
By definition  we have 
\begin{align}
\lambda_{f_\nu}(\x^*_\mu) \!=\! \frac{1}{\sqrt \nu} \| \nabla f_\nu(\x^*_\mu) \| \stackrel! \le \eta  
\iff \| \x^*_\mu\| \stackrel !\le \frac{\eta \sqrt{\nu}}{(\mu - \nu) }
\iff \nu^2 -   (2+B_\mu) \mu \nu + \mu^2 \stackrel !\le 0
\end{align}
Solving the quadratic equation for $\nu$ yields the claim.
\end{proof}
\end{lemma}

\begin{corollary}
Assume that $\phi(\mathbf 0,\z) \le \Phi$ ($\forall \z$).  Lemma \ref{lemma:erm-norm} remains valid with $B = \frac{\eta^2}{2\Phi} \ge B_\mu$.
\begin{proof} 
One can bound $\| \x^*_\mu\|$ easily through the following argument, exploiting non-negativity of $\phi$
\begin{align}
\Phi \ge f_0(\mathbf 0) = f_\mu (\mathbf 0)  \ge f_\mu(\x^*_\mu) = f_0(\x^*_\mu) + \frac{\mu}{2} \|\x^*_\mu\|^2
\ge \frac{\mu}{2} \|\x^*_\mu\|^2
\Longrightarrow \|\x^*_\mu\|^2 \le \frac{2 \Phi}{\mu} \,.
\end{align}
\end{proof}
\label{corollary:erm-lowermu}
\end{corollary}

Note that Corollary \ref{corollary:erm-lowermu} gives an \textit{a priori} rate guarantee, which only depends on the easily computable $\Phi$. This is a strong argument in favor of a continuation method, yet may not result in the most efficient schedules for reducing $\mu$. We will revisit this issue in Section~\ref{sec:data-adaptive-algorithm}.

\subsection{Increasing Sample Size}

We now generalize our analysis to the case, where we also increase the sample size. The basic challenge is that we need to upper bound the gradient norm contributions coming from new data points. Intuitively this depends on the generalization capability of the current iterate.
\begin{lemma} \label{lemma:expected-gradientnorm}
Assume $f$ has Lipschitz continuous gradients with constant $L$. Let $\S$ be a given $n$-sample, which we split into $\S_0 = (\z_1,\dots, \z_m)$ and $\S_1 = (\z_{m+1}, \dots,\z_n)$, where $m \le n$ is arbitrary.  Define $\x^* := \arg\min_\x f^{\S_0}(\x)$. With high probability over $\S_0$ it holds that 
\begin{align}
\E_{\S_1} \|  \nabla f^{\S}(\x^*) \|^2  \leq  \frac{2 L\left(n-m\right)}{n} \bound(m)
\end{align} 
\begin{proof} See Appendix. \end{proof}
\end{lemma}

\begin{theorem}
Under the same conditions as Lemma \ref{lemma:expected-gradientnorm} and for arbitrary $\nu \le \mu$ and $\beta \in (0;\infty)$,
\begin{align*}
\E_{\S_1} \| \nabla f^\S_{\nu}(\x^*)\|^2 \le &
 (1+\beta) \left( \nu -\mu  \right)^2  \| \x^*\|^2  + (1+\beta^{-1})  \left( \frac{n-m}{n}  \right)^3   2 L \bound(m) 
\end{align*}
\begin{proof}
See Appendix.
\end{proof}
\end{theorem}
The theorem directly implies that we can construct a sequence of objective functions with sample size increasing at a geometric rate. 

\begin{corollary}
There exists an $\alpha^* \in [0;1)$ such that for all $\alpha \in [\alpha^*; 1]$, with $m/\alpha \in {\mathbb Z}$, the following holds for $\nu := \alpha \mu$ and $\S=(\z_1,\dots,\z_{m/\alpha})$, 
\begin{align}
\| \nabla f^{\S}_{\nu}(\x^*) \| \le \eta \sqrt{m} \,,
\label{eq:alpha_existence}
\end{align}
where $\alpha^*$ can be explicitly computed as the root of a third order polynomial.
\begin{proof}
See Appendix. 
\end{proof}
\end{corollary}

\subsection{Data-Adaptive Algorithm}
\label{sec:data-adaptive-algorithm}

The above analysis is largely data-independent and just involves a few constants: $L$, $\Phi$, proportionality constants involving $\bound$ and $\mu \propto 1/m$. As such, it leads to geometric rates that can be quite conservative. We will now show a data-adaptive strategy that -- up to small approximation errors --  maximally increases the sample size, and, equivalently, maximally decreases the regularization strength, within the desired range. First of all, note that we can easily compute the gradient on an increased sample. Denote $\S= (\z_1,\dots,\z_n)$ and $\S_0 = (\z_1,\dots,\z_m)$, $m \le n$. Define $\x^* = \argmin_\x f^{\S_0}_{\mu}(\x)$, 
\begin{align}
\nabla f_{\nu}^\S(\x^*) = \frac{1}{n} \sum_{k=m+1}^n \nabla \phi^{z_k}(\x^*)  + \nu \left( 1 - \frac{\mu m}{\nu n} \right)  \x^* \,.
\end{align}
Note that if $\mu/\nu = n/m$, then the second term vanishes. As we need to compute the gradient anyway as part of the Newton iteration, we get it for free. We now could approximate 
\begin{align}
\lambda_{f^\S_{\nu}}(\x^*) \le \frac 1{\sqrt{\nu}} \| \nabla f_{\nu,n}(\x^*) \|\,,
\end{align}
but this results in bounds that are not very tight and hence underestimate the possible rate. We thus propose a tighter bound based on a Taylor approximation of the inverse Hessian at the previous iteration, something that we can compute with a little bit of extra computational effort.\footnote{Computing a Newton update is usually done via LU decomposition, which is cheaper.}  Let $\nu \le \mu$ and define $\mathbf H := \nabla^2 f^{\S_0}_{\mu}(\x^*)$ 
\begin{align}
\left[ \mathbf H - (\mu + \nu) \mathbf I \right]^{-1} =  \mathbf H^{-1} + (\mu - \nu) \mathbf H^{-2} + \mathbf O(\mu^2)
\end{align} 
and thus with the additional assumption that $\mathbf H \approx \nabla^2 f^\S_{\mu}(\x^*)$,
\begin{align}
\left[ \lambda_{f^\S_{\nu}}(\x^*) \right]^2\approx  \nabla f^\S_{\nu}(\x^*)^\top \mathbf H^{-1} \nabla f^\S_{\nu}(\x^*) + (\mu-\nu) \| \mathbf H^{-1} \nabla f^\S_{\nu}(\x^*)\|^2\,.
\end{align}
For instance, in the setting $\nu \propto 1/n$, it is straightforward to (numerically) find the maximal $n$ -- or equivalently minimal $\nu$ -- such that the above approximation is $ < \eta^2$. In our experiments, we have found this approximation to be correct up to a few percent relative error. 

\begin{algorithm}[t]
\begin{algorithmic}[1]
\STATE Given sample $\S$ and $(\mu_0,m_0$), define $\S_0 := \S_{1:m_0}$
\STATE $\x_0 \leftarrow \argmin f^{\S_0}_{\mu_0}$ 
\FOR{$t=1,\dots,T$}
\STATE Find the smallest $\alpha$ such that 
\vspace{-2mm}
\begin{align*}
\lambda_{f^\alpha}(\x^*) \le \eta, \quad \text{where} \quad f^\alpha := f^{\S'}_{\nu}, \; \nu := \alpha \mu, \; \S':= \S_{1:n}, \; n:=m_{t-1}/\alpha.
\end{align*}
\vspace{-0.7cm}
\STATE $(\mu_t,m_t )\leftarrow (\nu, n)$
\STATE Compute Newton increment $\triangle \x$ for $f^\alpha$ at $\x_{t-1}$
\STATE $\x_t \leftarrow \x_{t-1} + \triangle \x$  
\ENDFOR
\end{algorithmic}
\caption{Data-adaptive Newton's method - {\sc DynaNewton}}
\label{alg:data_adaptive_newton}
\end{algorithm}

The complete algorithm is summarized in Algorithm~\ref{alg:data_adaptive_newton}. In order to find an $\alpha$ that fulfils Eq.~\eqref{eq:alpha_existence_2}, we need to compute an estimate of $\left\| \nabla f^\S_{\mu}(\x^*)\right\|$. We solve this problem by first assuming that $\x_t$ is a good approximation of the solution to $f_t$ and then computing the estimate from the samples $(\z_{m+1},\dots,\z_{m+k})$. We can avoid the first approximation at the cost of making $5-6$ Newton iterations instead of just $1$, but we have found this not to be necessary in any one of our  experiments. Further experimental results are provided in the appendix.

\section{Experiments}

\paragraph{Datasets}
We apply $\ell_2$-regularized logistic regression on a set of 4 datasets of varying size and dimensionality summarized in Table~\ref{table:datasets}. We use $90\%$ of the data points as the training set and the remaining $10\%$ as test set.
 
\begin{table}[h!]
\begin{center}
\begin{small}
\begin{sc}
\begin{tabular}{lcc} 
\hline
Dataset & Size & Number of features 
\\
\hline
a9a & 32561 & 123 \\
w8a & 49749 & 300 \\
covtype.binary & 581012 & 54 \\
SUSY & 5000000 & 18
\\
\hline
\end{tabular}
\end{sc}
\end{small}
\end{center}
\caption{{\it Details of the datasets used in our experiments.}}
\label{table:datasets}
\vspace{-3mm}
\end{table}
 
\begin{figure*}
	\begin{center}
          \begin{tabular}{@{\hspace{5mm}}c@{\hspace{5mm}}c}
            \includegraphics[
            width=0.40\linewidth]{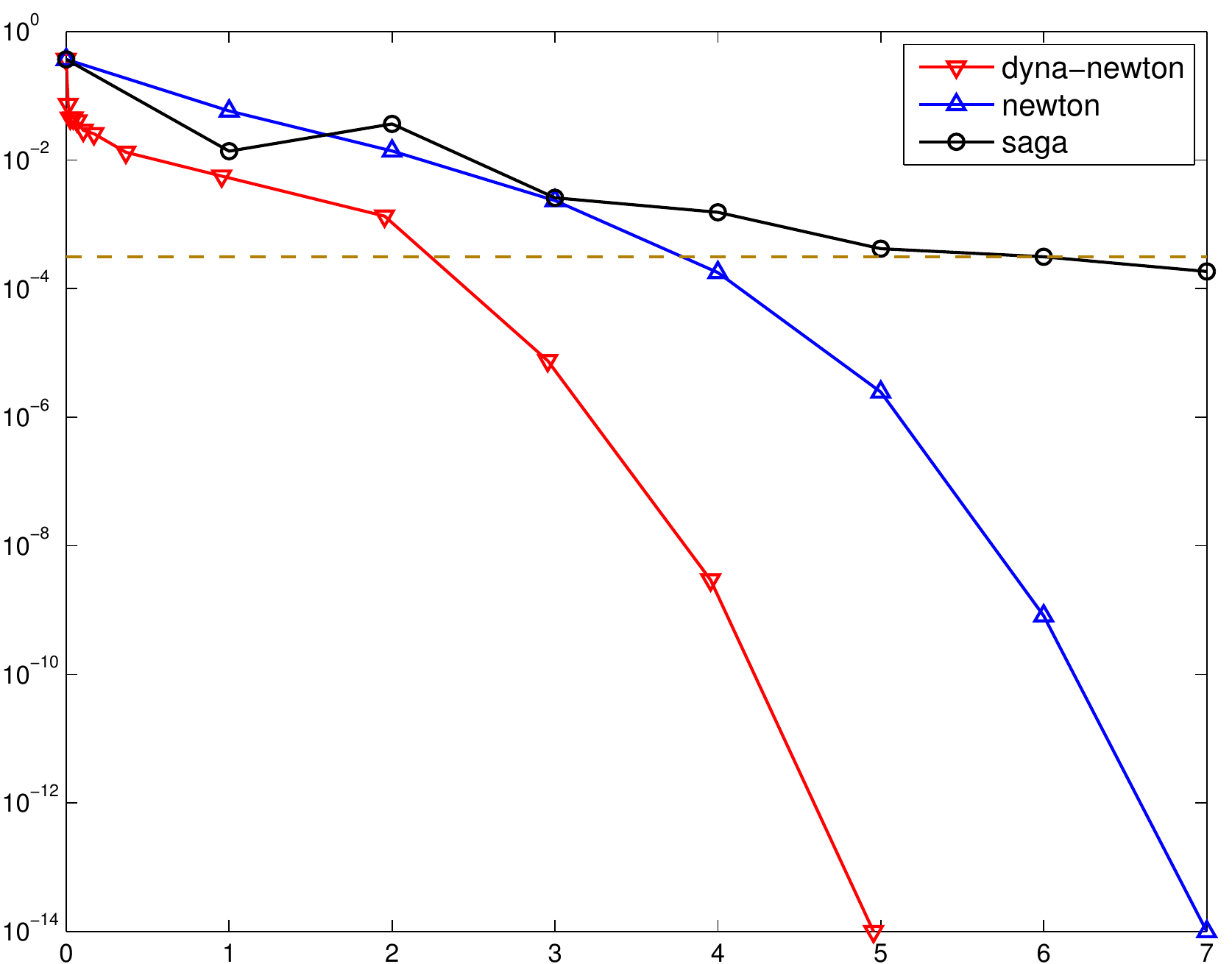} & 
            \includegraphics[
            width=0.40\linewidth]{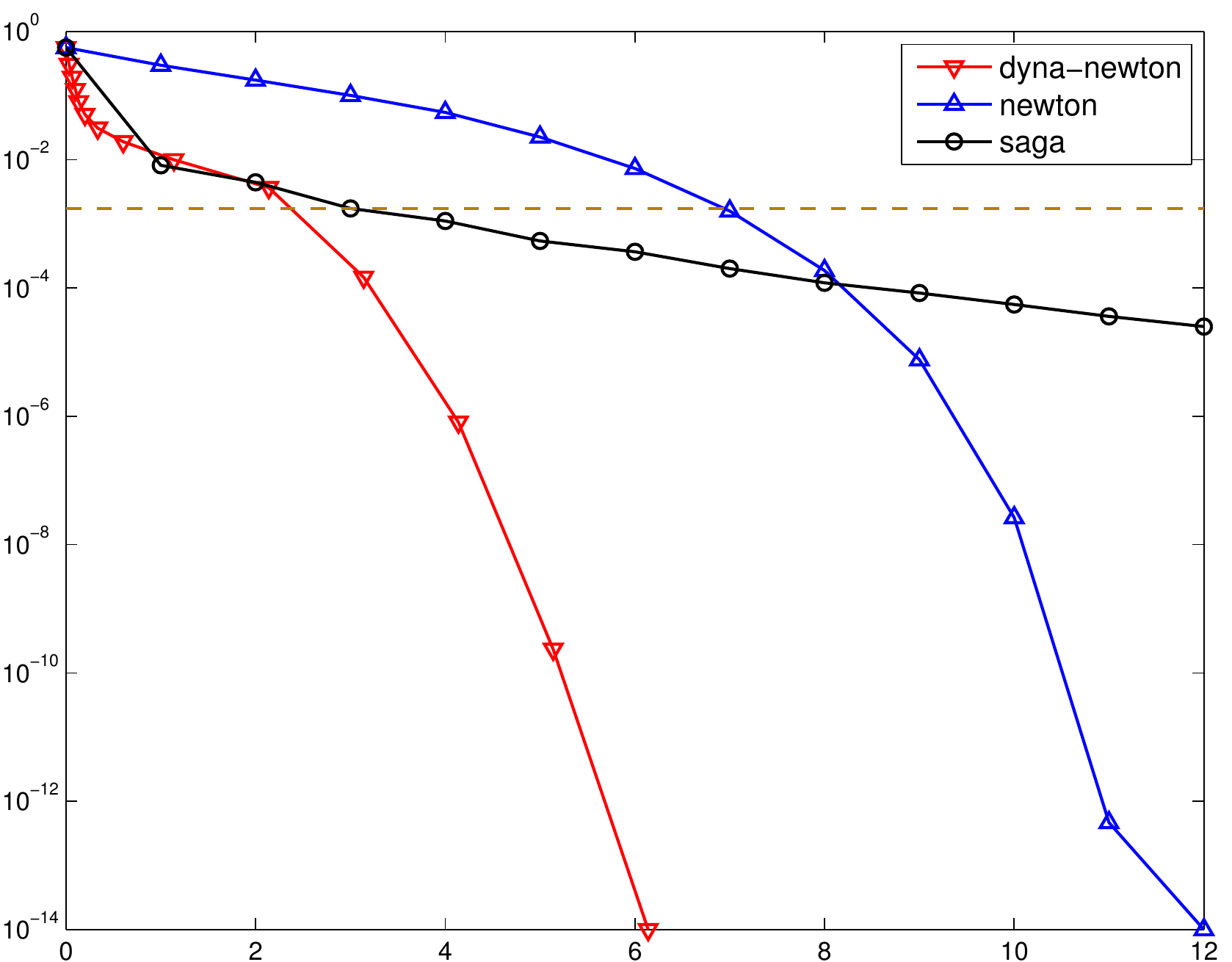}  \\
                1.  {\sc a9a}       &
                 2. {\sc w8a}  
                \\
               \includegraphics[width=0.40\linewidth]{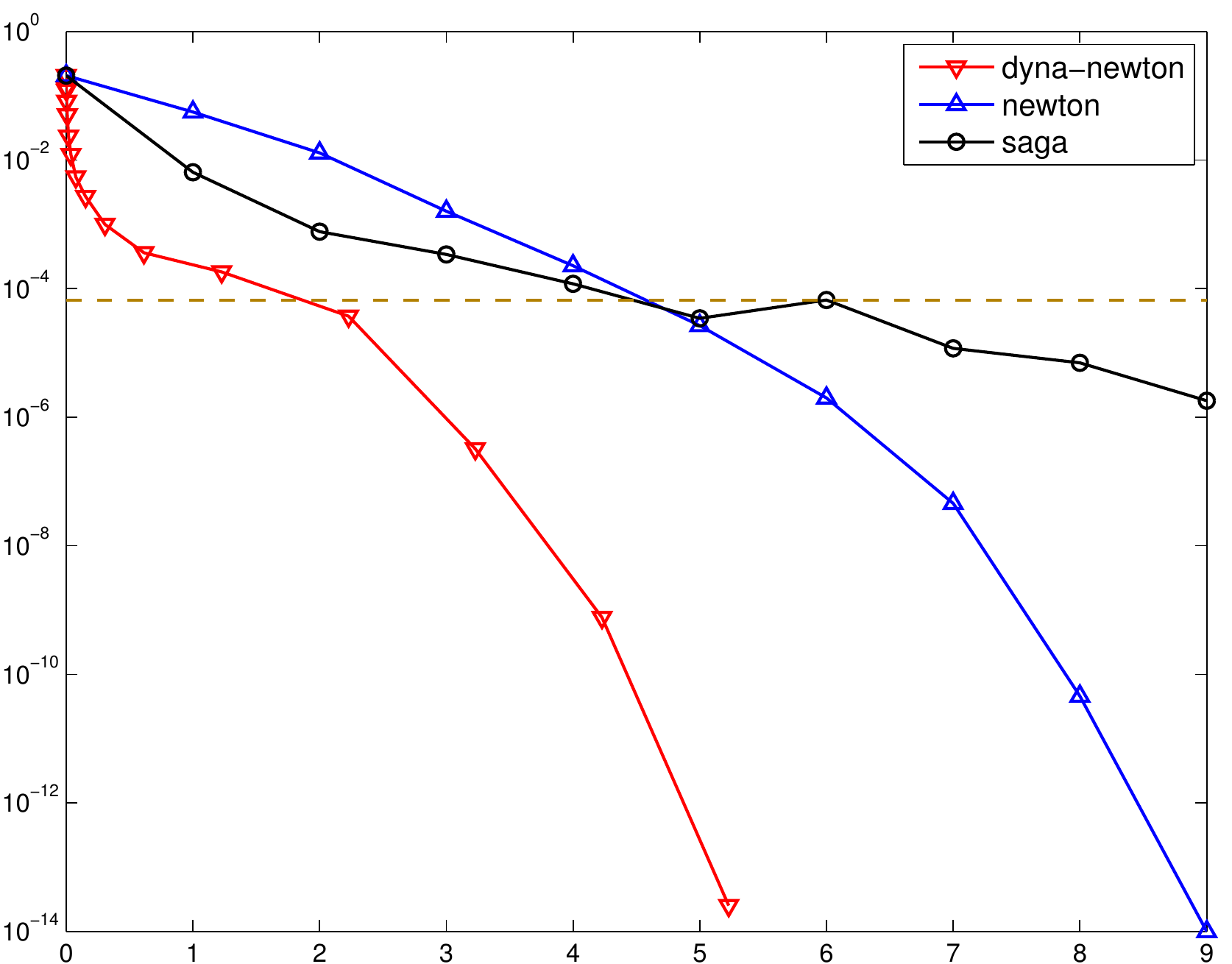} &
            \includegraphics[width=0.40\linewidth]{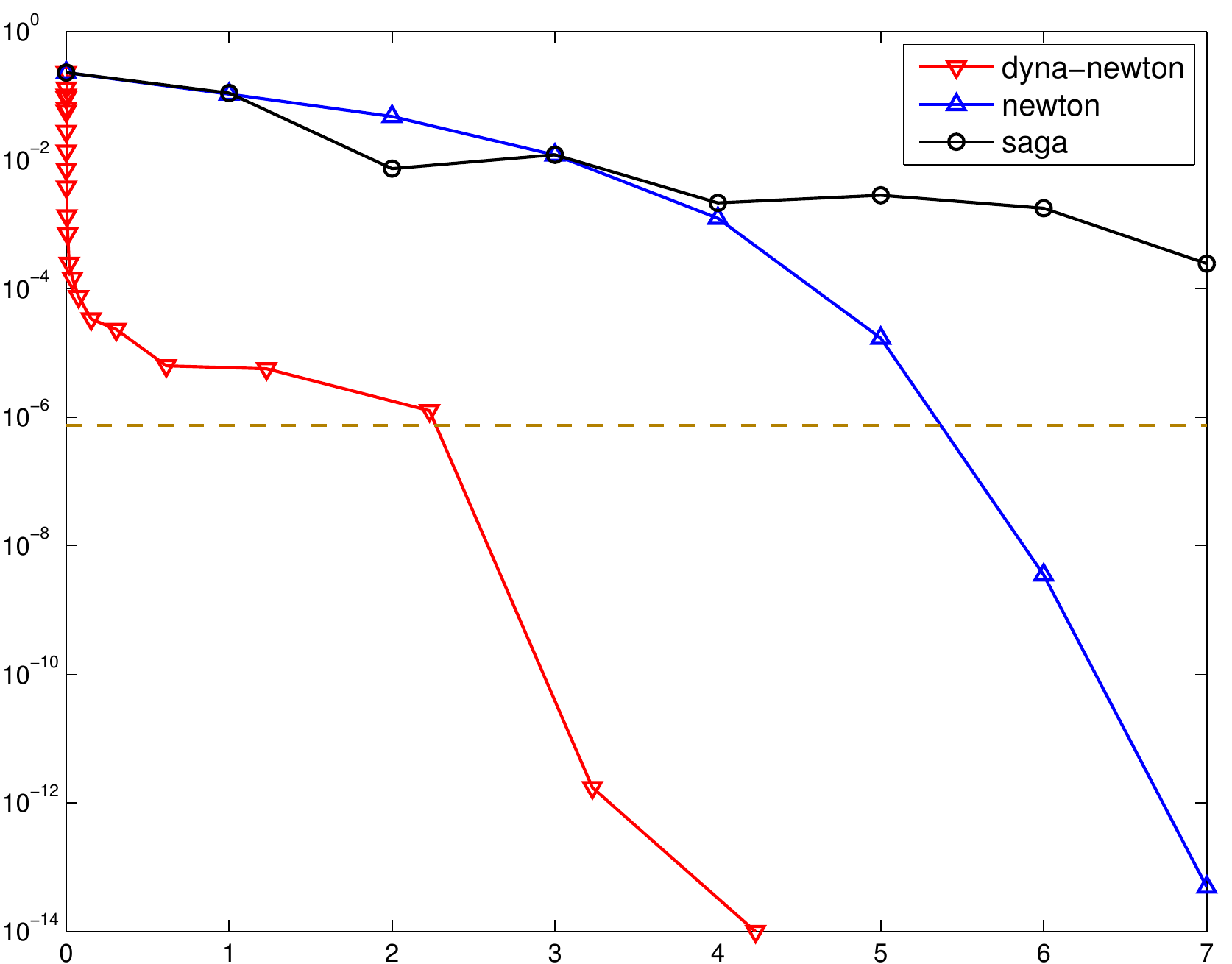}
            \\ 
          3. {\sc covtype} &
            4. {\sc susy} 
	  \end{tabular}
          \caption{{\it Suboptimality on the empirical risk vs effective number of epochs}.~The
          plots show the suboptimality of the empirical risk on the full data set $\S^*$ with $\nu = 1/N$. The horizontal axis represents the number of {\it effective epochs}, i.e. number of passes over the whole training data set. 
          The dotted horizontal line refers to statistical accuracy (explained in the main text). }
          \label{fig:results_baselines_epochs}
	\end{center}
\end{figure*}

\paragraph{Comparison to standard baselines} We compare {\sc DynaNewton} (cf.~Algorithm~\ref{alg:data_adaptive_newton}) to Newton's method and -- as a competitive SGD variant -- to SAGA. We used a step size of $\sim \frac1L$ for SAGA as suggested in previous work~\cite{defazio2014saga, hofmann2015variance}. The results presented in Figure~\ref{fig:results_baselines_epochs} show significant speeds-ups compared to Newton's method. {\sc DynaNewton} also outperforms SAGA and gets a very accurate solution after less than 6 epochs on all datasets. We also evaluated the solution quality on the expected risk and provide the complete results of this evaluation in the Appendix (see Figure~\ref{fig:results_expected_epochs}). In order to relate convergence on the empirical and expected risks, we plotted a horizontal dotted line that shows the iteration at which {\sc DynaNewton} reached convergence on the test set. This demonstrates that {\sc DynaNewton} also achieves significant gains in terms of test error.

As the cost per iteration is typically higher for Newton's method on a single machine, we also present a comparison in terms of running time in the Appendix (see Figure~\ref{fig:results_baselines_time}). Although the gains are more moderate when measuring time, it is worth pointing out that Newton is inherently much easier to parallelize as discussed in~\cite{dean2012large} and further gains are thus to be expected in a distributed setting.
 
\begin{figure*}
	\begin{center}
            \includegraphics[width=0.45\linewidth]{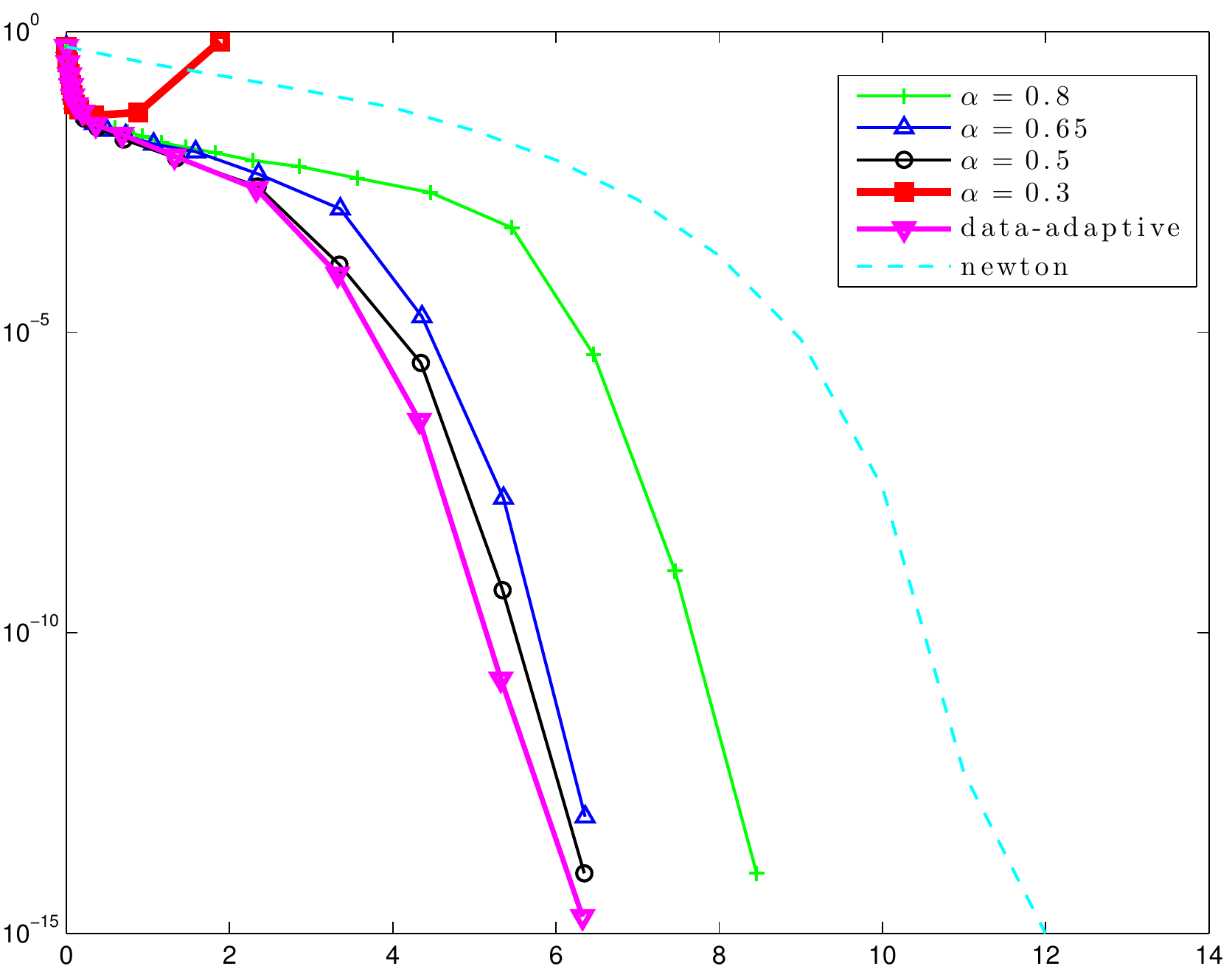} 
           \caption{{\it Increment Factor test on the {\sc w8a} dataset}.~The
          plots shows the suboptimality of the empirical risk on the full data set $\S^*$ with $\nu = 1/N$. In these experiments we used the increment factor $\alpha = m_{t-1}/m_t$. It is clear that the smallest $\alpha$ value causes leaving the region of quadratic convergence and hence leads to divergence. This behavior is excluded by the data-adaptive approach.}
	\end{center}
	\label{fig:increment_factor}
\end{figure*}

\paragraph{Increment factor test}
We evaluate the influence of $\alpha$ on the convergence of the algorithm without data-adaptivity. As can be seen in Figure~\ref{fig:increment_factor} a small value of $\alpha$ leads to faster convergence while it might also make the algorithm diverge if chosen too aggressively small. In contrast, the data-adaptive approach adapts the value of $\alpha$ and yields a stable convergence behavior on all datasets. We observed empirically that the data-adaptive method chooses a value of $\alpha \approx \frac12$ thus explaining why the non-adaptive approach with $\alpha=\frac12$ achieves a similar - but slightly inferior - performance.

\begin{figure*}
	\begin{center}
          \begin{tabular}{@{\hspace{5mm}}c@{\hspace{5mm}}c}
            \includegraphics[
            width=0.40\linewidth]{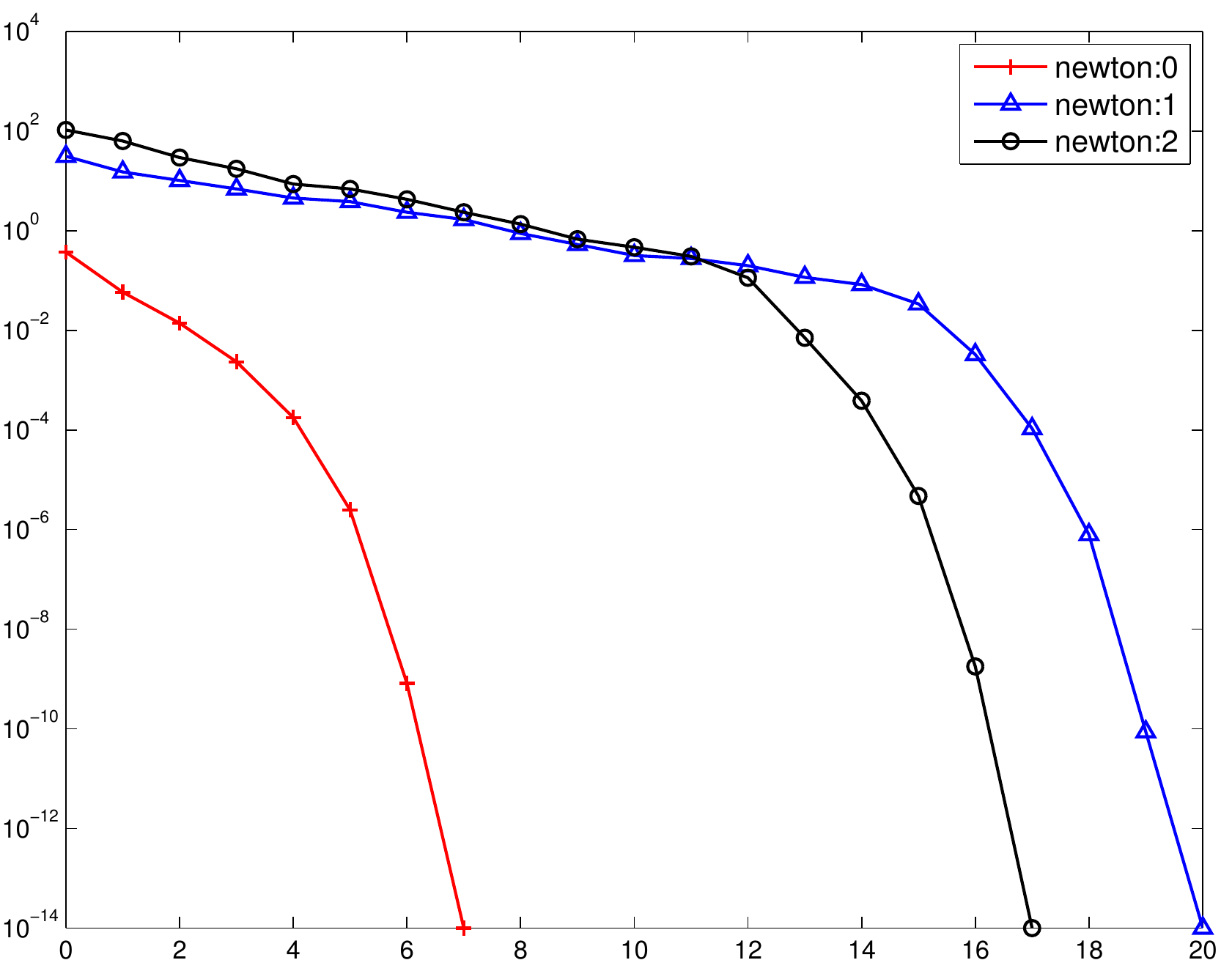} & 
            \includegraphics[
            width=0.40\linewidth]{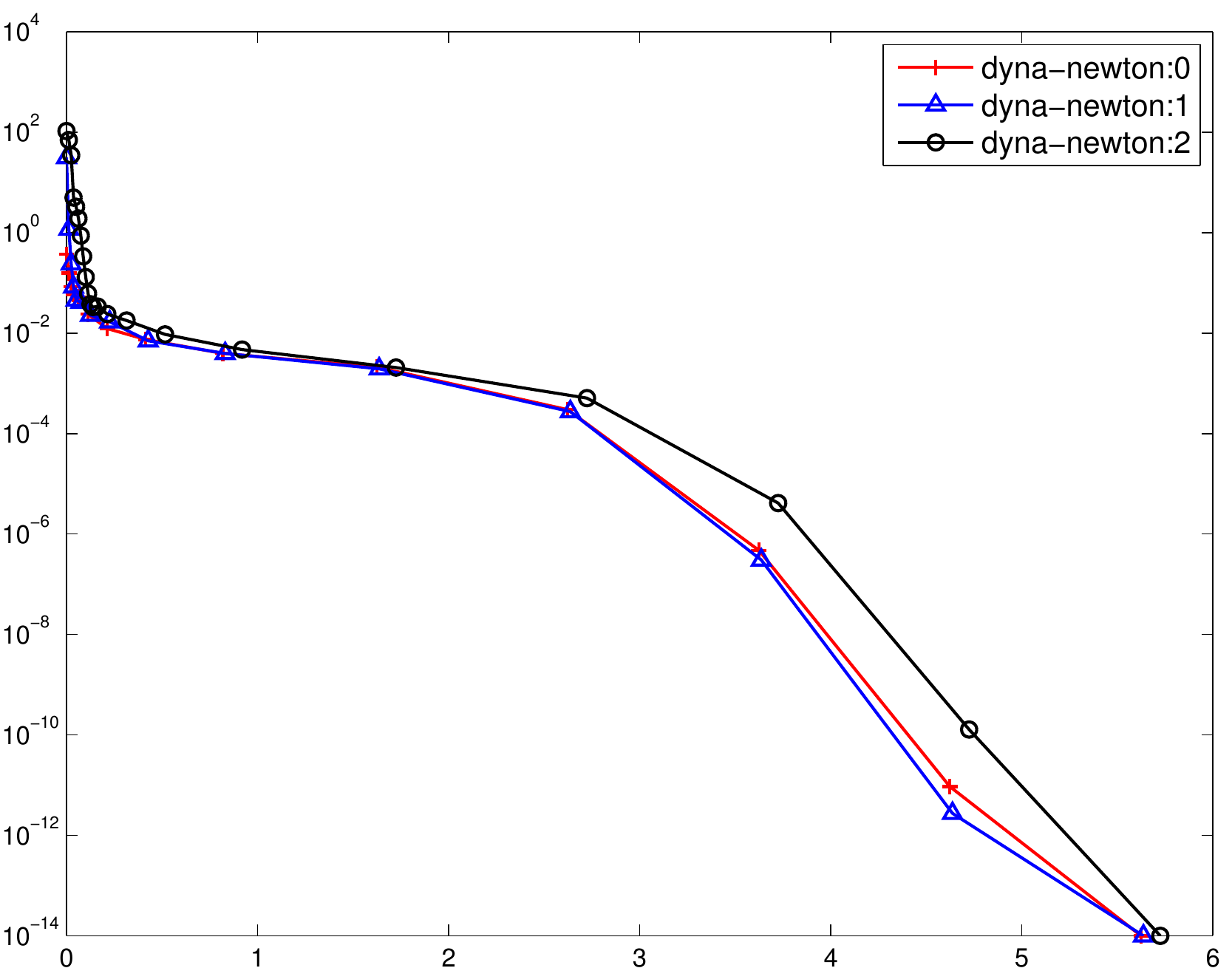}  \\
                1.  {\sc newton}       &
                2. {\sc dyna-newton}  
	  \end{tabular}
          \caption{{\it Robustness against the initialisation point on the {\sc a9a} dataset}. We here show the sub-optimality on empirical risk against time for various initialization points: in red, $\x_0 = \vec{0}$, in blue $\x_0=\vec{3}$ and in black $\x_0 = \vec{10}$.           }
          \label{fig:initialization_point}
	\end{center}
\end{figure*}

\paragraph{Importance of the initialization point}
We investigate the role of the initialization point on the convergence of Newton's method and {\sc DynaNewton}. The results shown in Figure~\ref{fig:initialization_point} demonstrate that bad initialization points (i.e.~far away from the optimum) significantly slow down the convergence of Newton's method as they require more steps to get inside the ball of quadratic convergence. By comparison, a poor initialization does not significantly affect {\sc DynaNewton}.

\section{Conclusion}

We proposed a continuation method variant of Newton's method that dynamically  adapts the sample size and the regularization strength such that each subproblem provides a starting point that is within the quadratic convergence region of the subsequent optimization problem. We provided a theoretical analysis that characterizes the conditions required for achieving a proper hand-over and also devised a data-adaptive strategy of discretizing the solution path. 

Our empirical results demonstrate significant speed-ups across a wide range of datasets both in terms of empirical as well as expected risk. In particular the speed of convergence on the latter is quite remarkable, often getting close to optimal solutions in about $2$ effective epochs. All results seem to be in good agreement with our theory and its predictions. 

In the appendix, we provide further empirical evidence that shows that our results extend to non-exact Newton methods such as L-BFGS. This is of special interest for training large deep networks for which a distributed implementation of L-BFGS has been shown to provide significant speed-ups in terms of training time~\cite{dean2012large}. While our analysis does not provide any guarantees for this case, it seems that the proposed continuation method has merits beyond the results presented in this paper. A better understanding of the behavior of quasi-Newton methods is the topic of  future work.

\subsubsection*{Acknowledgments}
We would like to thank Aryan Mokhtari for helpful discussions on Newton's method.

\bibliography{arxiv_newton_full}
\bibliographystyle{abbrv}

\newpage
\appendix
\section{Appendix}

\subsection{Proofs}

\begin{LemmaM}{\bf 3}
\begin{proof}
Using classical concentration inequality from statistical learning theory we get (see also \cite{daneshmand2016small}) with high probability over $\S_0$ 
\begin{align} \label{eq:switching_bound}
\E_{\S_1} \left[ f^\S(\x^*)  - \min_\x f^\S(\x) \right] \leq \frac{n-m}{n} \bound(m) \,,
\end{align}
which bounds the suboptimality of the minimizer of the smaller $m$-sample on the larger $n$-sample.  Furthermore, by virtue of smoothness:
\begin{align}
\|  \nabla f^\S_{}(\x^*) \|^2  & \leq  2 L \left[ f^\S_{}(\x^*)  - \min_\x f^\S_{}(\x) \right] 
\end{align}
Putting both inequalities together yields the claim.
\end{proof}
\end{LemmaM}

\begin{TheoremM}{\bf 4}

\begin{proof}
\begin{align}
\nabla f^\S_{\nu}(\x^*) & = 
\frac{m}{n} \nabla f^{\S_0}_{0} (\x^*)  + 
\frac{n-m}{n} \nabla f^{\S_1}_{0}(\x^*) + 
\nu \x^*
\\ \nonumber
& = -\frac mn \mu \x^* + \frac{n-m}{n}  \nabla f^{\S_1}_{\mu}(\x^*)
- \frac{n-m}{n} \mu \x^* + \nu \x^*
\\ \nonumber 
& = \left( \nu - \mu \right) \x^* + \frac{n-m}{n}  \nabla f^{\S_1}_{\mu}(\x^*) 
 =  \left( \nu - \mu \right) \x^* + \frac{n-m}{n}  \nabla f^\S_{\mu}(\x^*)  \,.
\end{align}
Here we have repeatedly exploited  the first order optimality condition of $\x^*$, i.e.~$\nabla f^{\S_1}_{\mu}(\x^*) =0$. 
 Now, taking squared norms on both sides, we apply the generalized parallelogram identity  $\| a + b \|^2 \le (1+\beta) \| a\|^2 + (1+ \beta^{-1}) \| b\|^2$ with $\beta \in (0; \infty)$. 
\begin{align}
\| \nabla f^\S_{\nu}(\x^*)\|^2 
\le \;\; &  (1+\beta) \left( \nu -  \mu  \right)^2  \| \x^*\|^2 + (1+\beta^{-1})  \left( \frac{n-m}{n}  \right)^2 \left\| \nabla f^\S_{\mu}(\x^*)\right\|^2 
\end{align}
Taking expectation with regard to $\S'$ on both sides and  applying Lemma~\ref{lemma:expected-gradientnorm} concludes the proof. 
\end{proof}
\end{TheoremM}

\begin{CorollaryM}{\bf 5} 
\begin{proof}
For concreteness set $\beta=1$. We want to chose $\alpha$ such that 
\begin{align}
& \mu^2  \| \x^*\|^2  (1-\alpha)^2 +  2 L \bound(m)  \left( 1-\alpha \right)^3  
\le \frac{\mu \eta^2}{2}
\label{eq:alpha_existence_2}
\end{align}
For $\alpha=1$, the left hand side is $0$, while the right hand side is strictly positive. As the derivative of the left hand side is finite at $1$, we can solve for $\alpha<1$ (such that $\alpha \ge 0$) as claimed. 
\end{proof}
\end{CorollaryM}

\subsection{Running time}

\begin{figure*}[h]
	\begin{center}
          \begin{tabular}{@{\hspace{5mm}}c@{\hspace{5mm}}c}
            \includegraphics[
            width=0.45\linewidth]{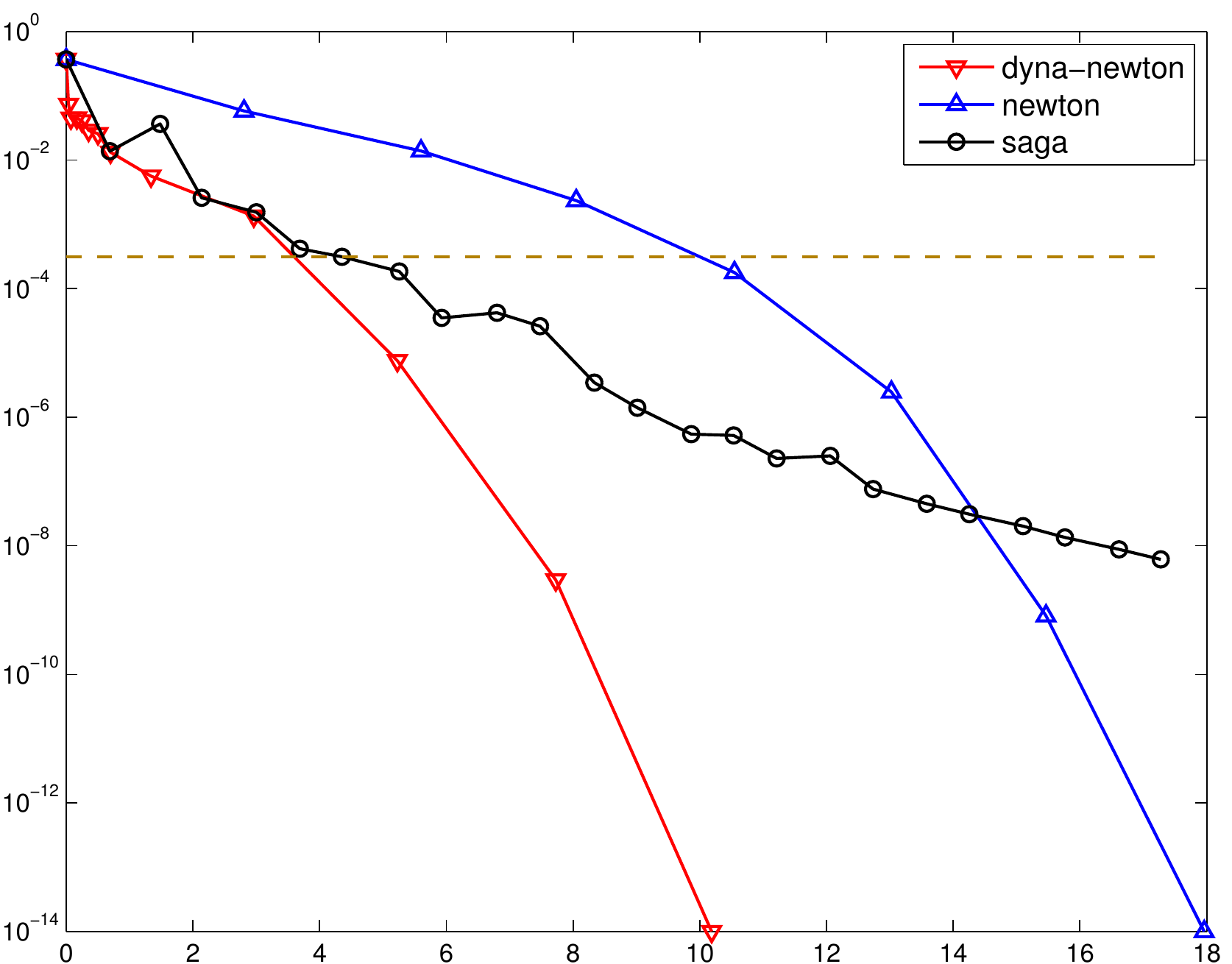} & 
            \includegraphics[
            width=0.45\linewidth]{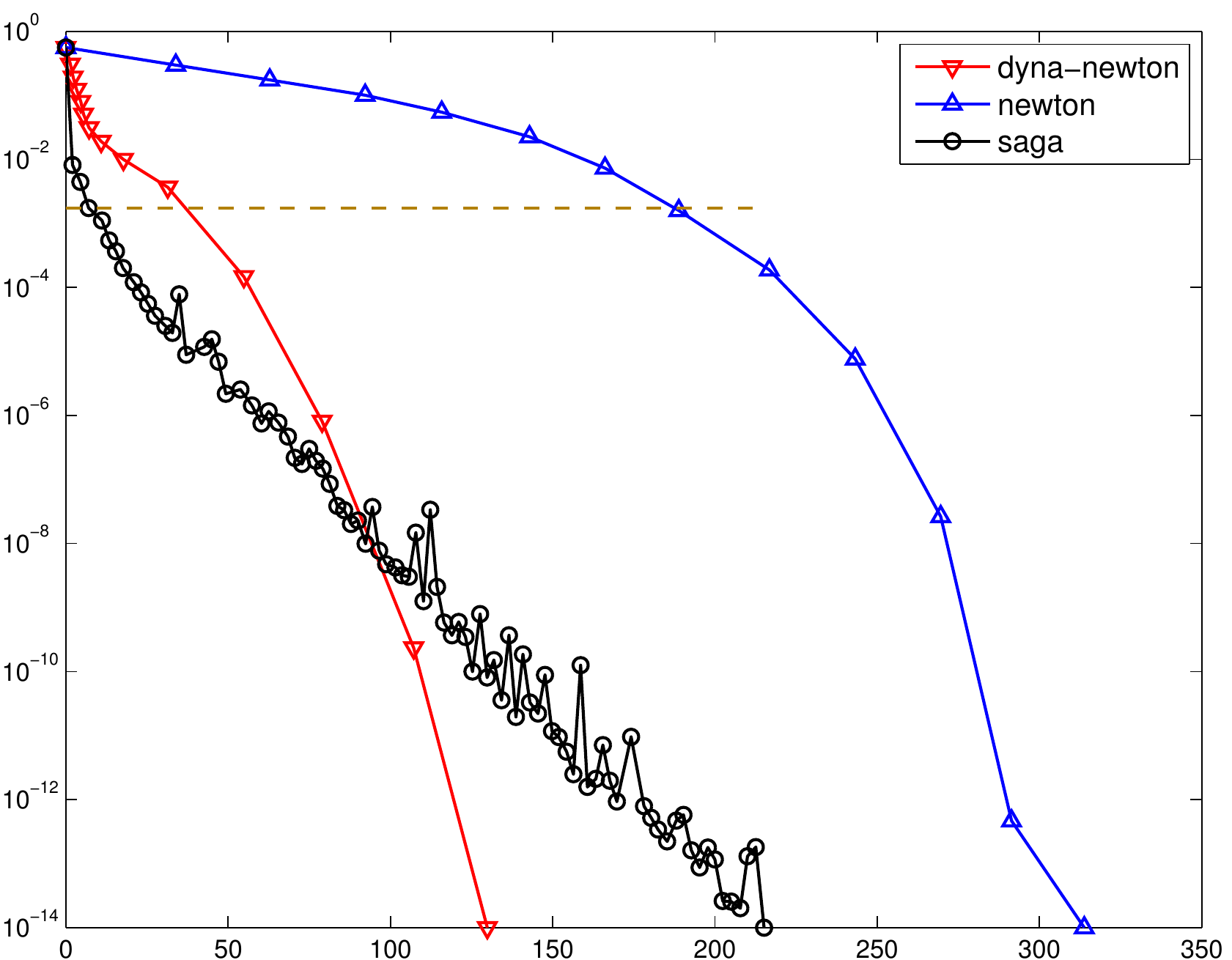}  \\
                1.  {\sc a9a}       &
                 2. {\sc w8a}  
                 \\
            \includegraphics[width=0.45\linewidth]{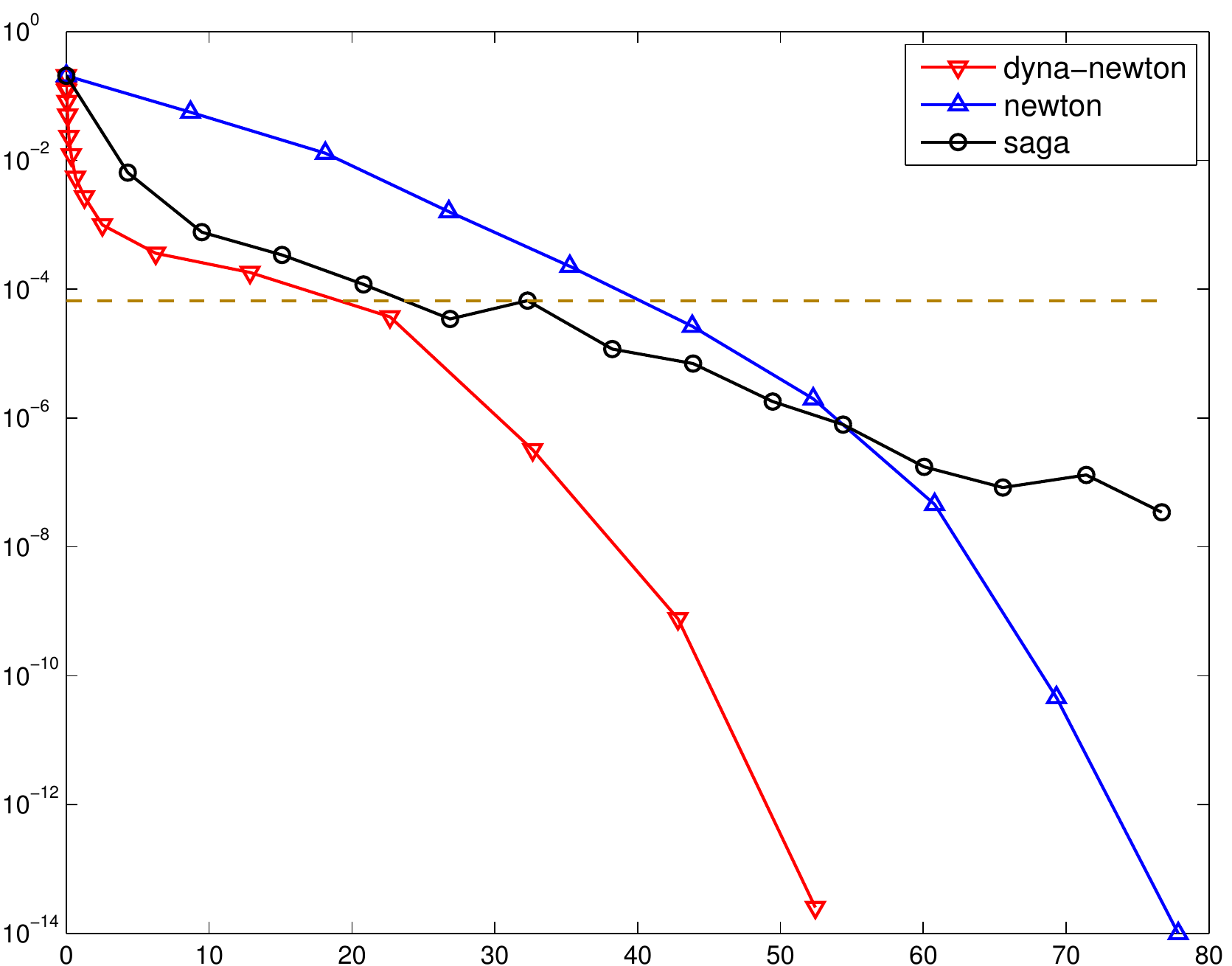} &
            \includegraphics[width=0.45\linewidth]{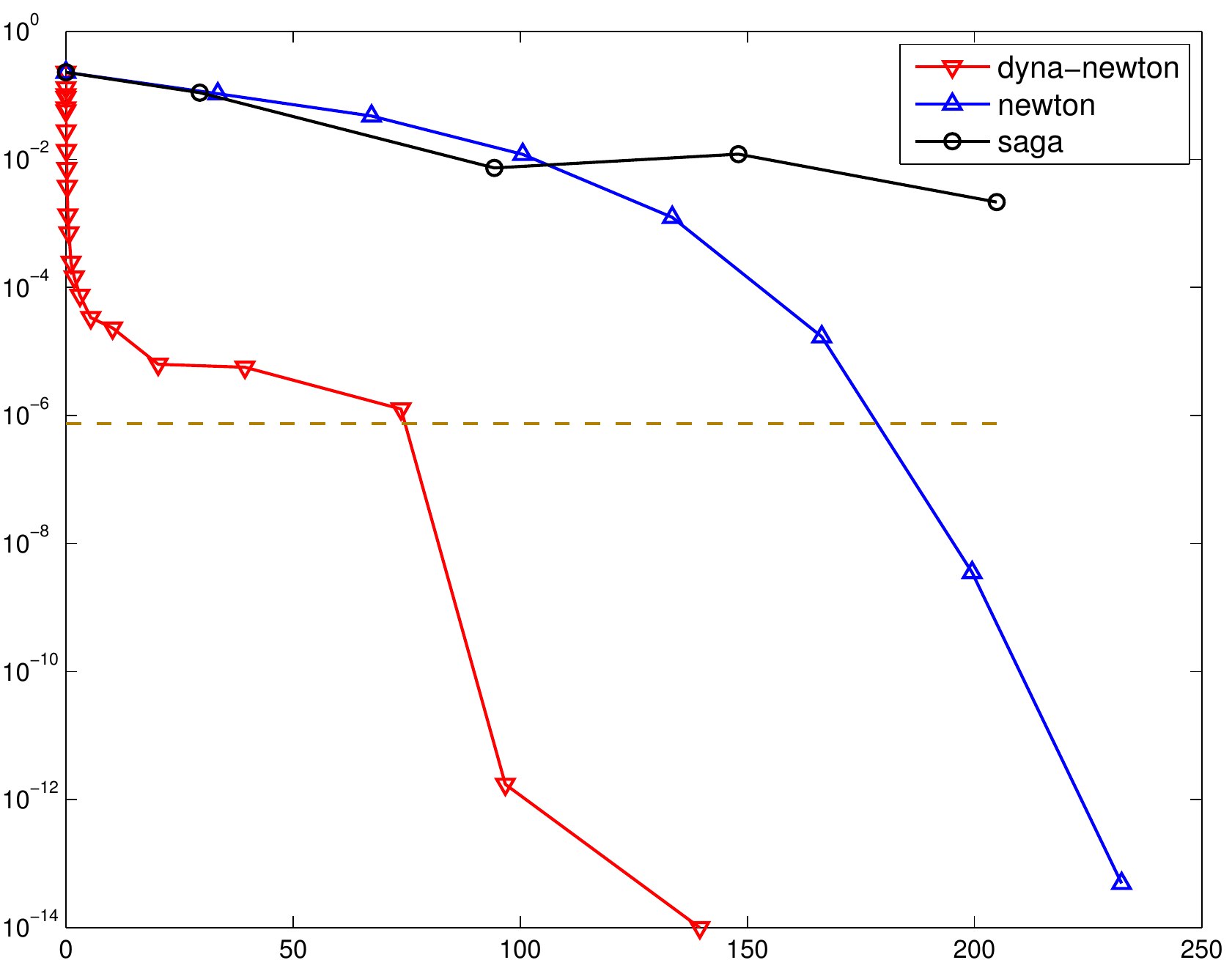}
            \\ 
            3. {\sc covtype} &
            4. {\sc susy} 
	  \end{tabular}
         \caption{{\it Suboptimality on the empirical risk vs time}.~The
          vertical axis shows the suboptimality of the empirical risk, i.e. $
         f_{\nu,n}(\x^t) -  f_{\nu,n}^*$,
          where $\nu = 1/n$. The horizontal axis represents run time in seconds. The horizontal dotted line corresponds to the iteration at which convergence is reached on the expected risk (see details in the main text).}
          \label{fig:results_baselines_time}
	\end{center}
\end{figure*}

We compare the running time of {\sc DynaNewton} against other baselines in Figure~\ref{fig:results_baselines_time}. Although we see more moderate gains in comparison to SAGA in terms of computational time (especially in early iterations), it is worth pointing out that Newton is inherently much easier to parallelize as discussed in~\cite{dean2012large} and further gains are thus to be expected in a distributed setting.

\subsection{Expected risk}

\begin{figure*}
	\begin{center}
          \begin{tabular}{@{\hspace{5mm}}c@{\hspace{5mm}}c}
            \includegraphics[
            width=0.40\linewidth]{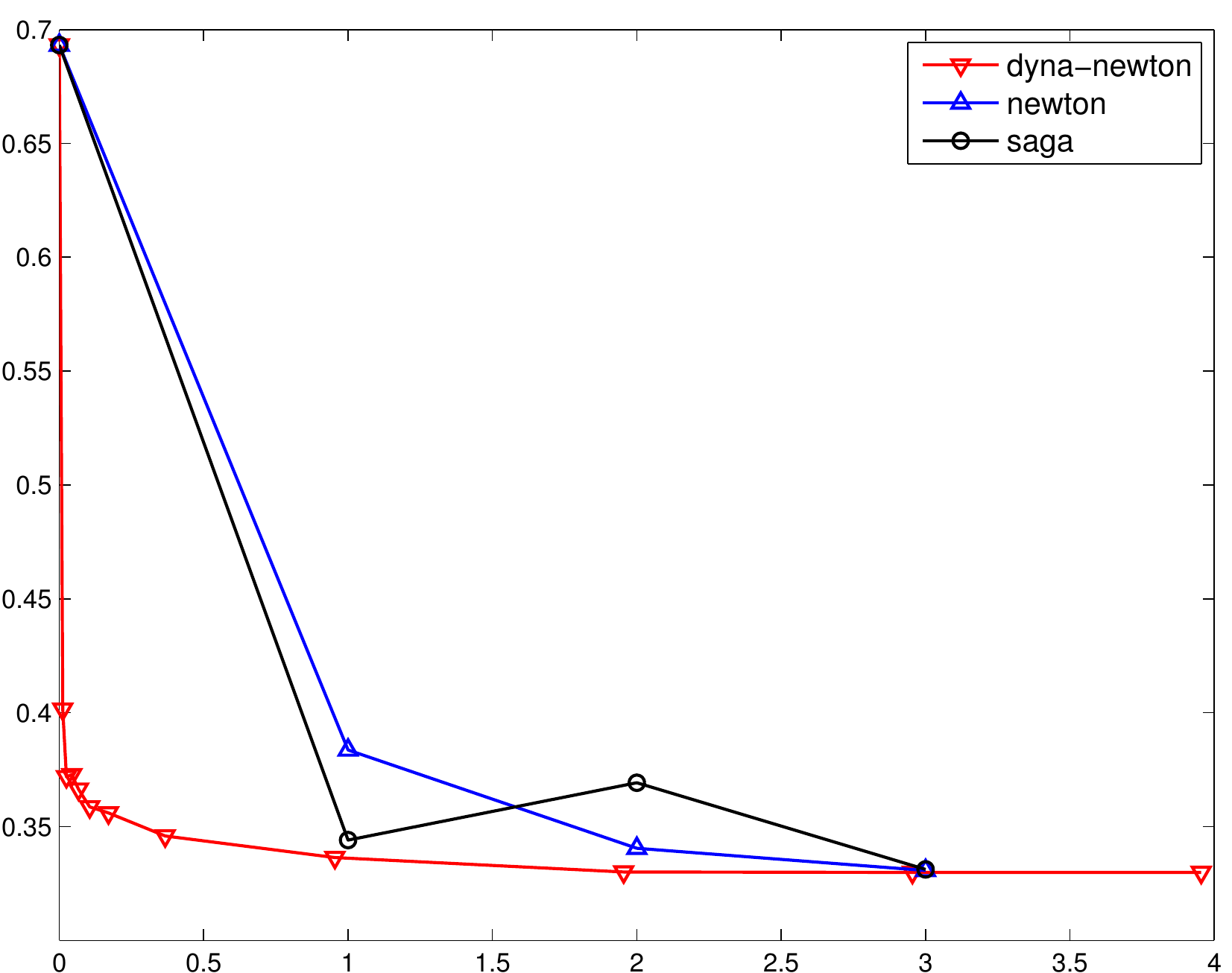} & 
            \includegraphics[
            width=0.40\linewidth]{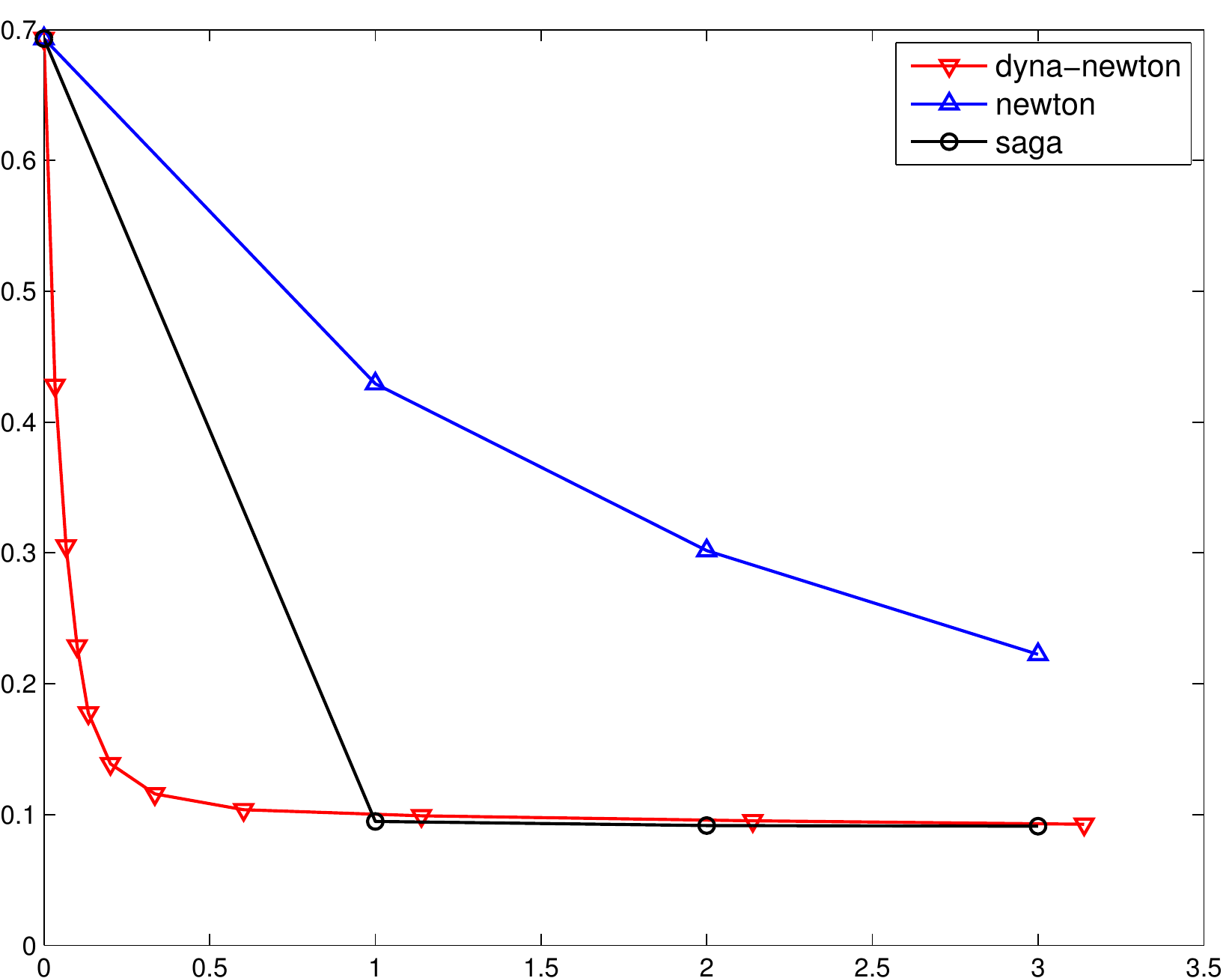}  \\
                1.  {\sc a9a}       &
                 2. {\sc w8a}  
                \\
               \includegraphics[width=0.40\linewidth]{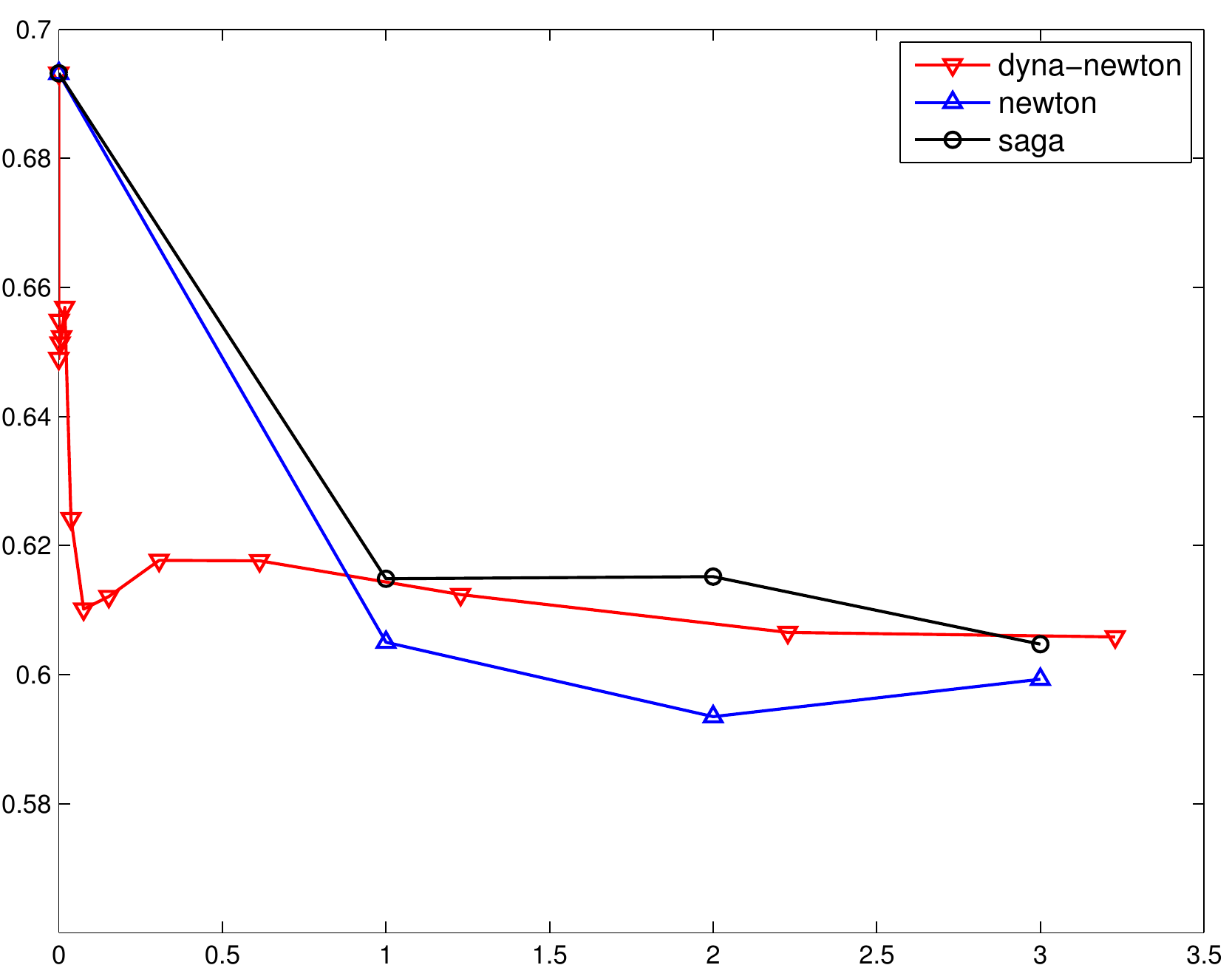} &
            \includegraphics[width=0.40\linewidth]{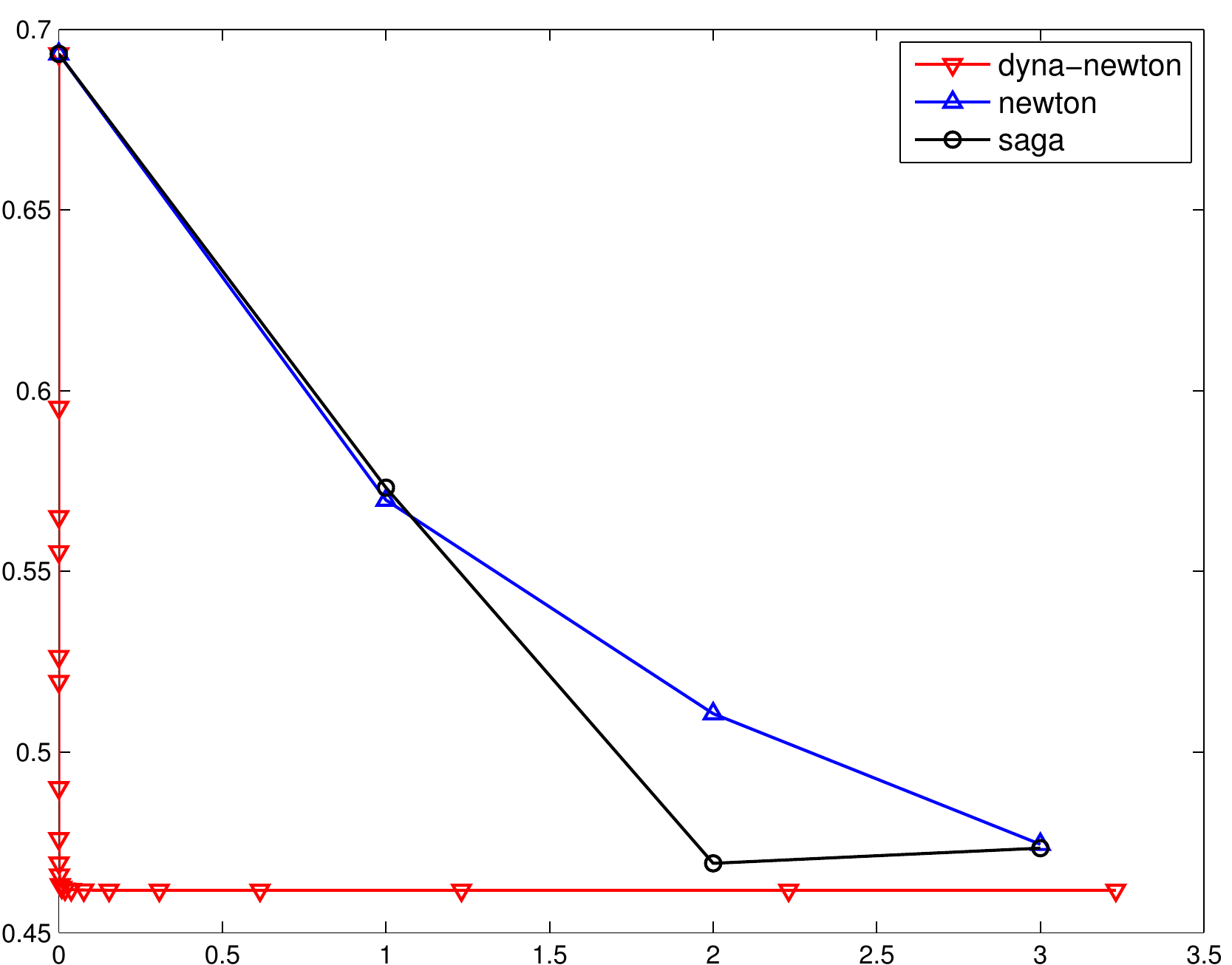}
            \\ 
          3. {\sc covtype} &
            4. {\sc susy} 
	  \end{tabular}
          \caption{{\it Suboptimality on the test set vs number of epochs}.~The
          vertical axis shows the suboptimality of the expected risk, i.e. $\phi_\T(\x) =  \sum_{\z \in \T} \phi_\z(\x) /|\T|$,
          where $\T$ is the test set. The horizontal axis represents the number of epochs.}
          \label{fig:results_expected_epochs}
	\end{center}
\end{figure*}

We present the results in terms of expected error in Figure~\ref{fig:results_expected_epochs}. This largely confirms the results obtained on the training set. Although all methods achieve convergence after less than 3 epochs, {\sc DynaNewton} achieves even faster convergence than Newton and SAGA.

\subsection{Approximation}

\begin{figure*}
	\begin{center}
          \begin{tabular}{@{\hspace{5mm}}c@{\hspace{5mm}}c}
            \includegraphics[
            width=0.45\linewidth]{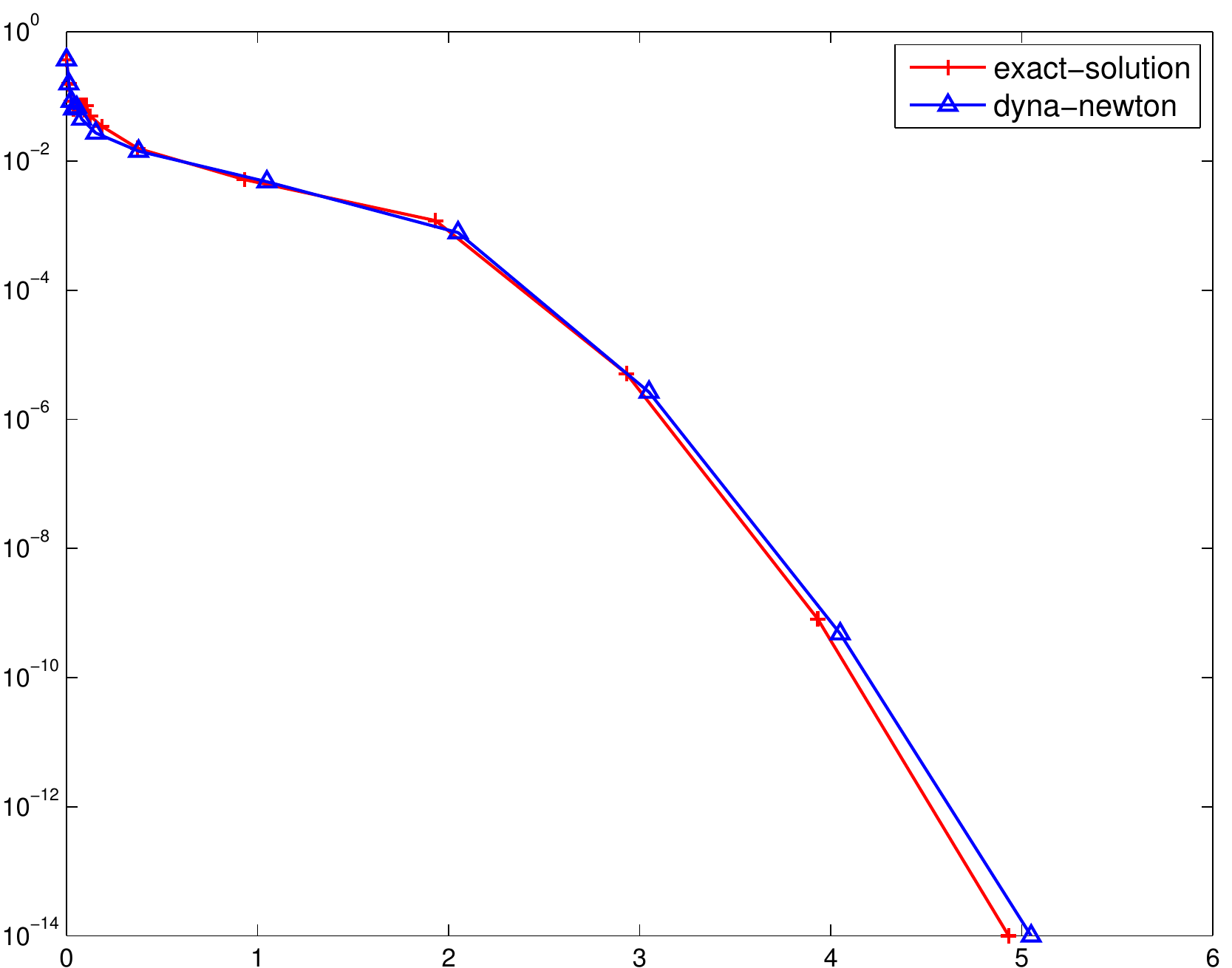} & 
            \includegraphics[
            width=0.45\linewidth]{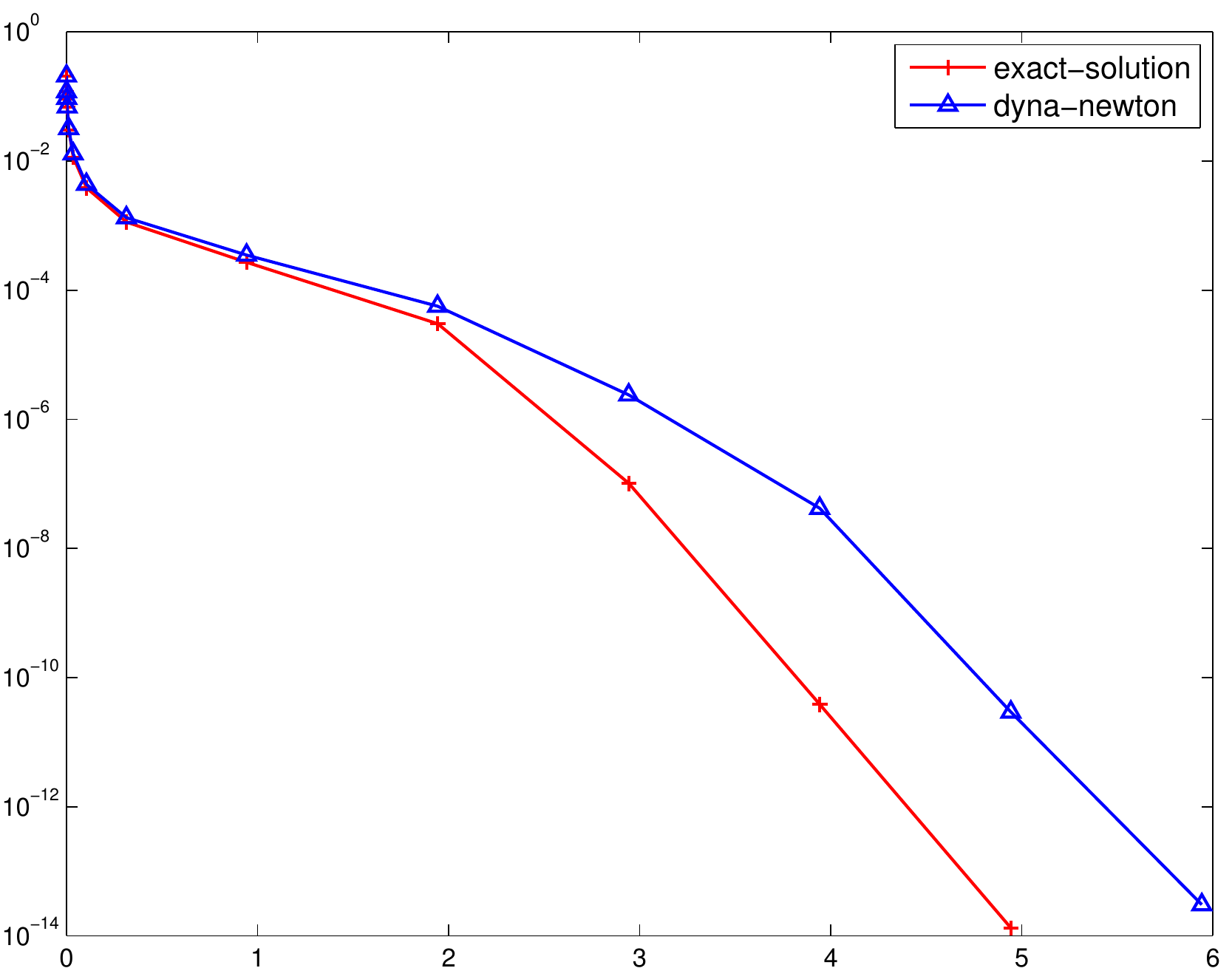}  \\
                1. Empirical suboptimality on {\sc a9a}       &
                 2. Empirical suboptimality on {\sc covtype}  
                \\
               \includegraphics[width=0.45\linewidth]{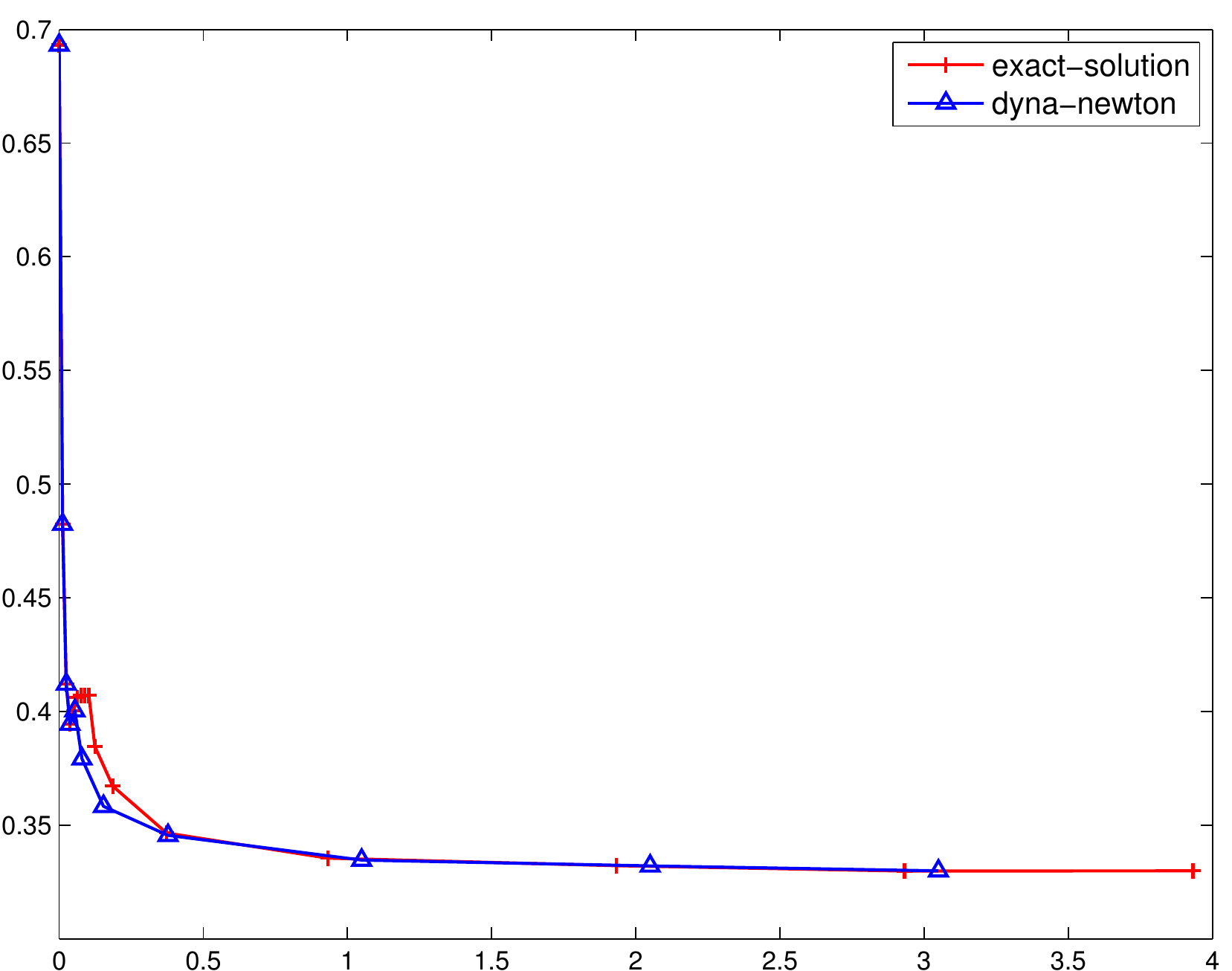} &
            \includegraphics[width=0.45\linewidth]{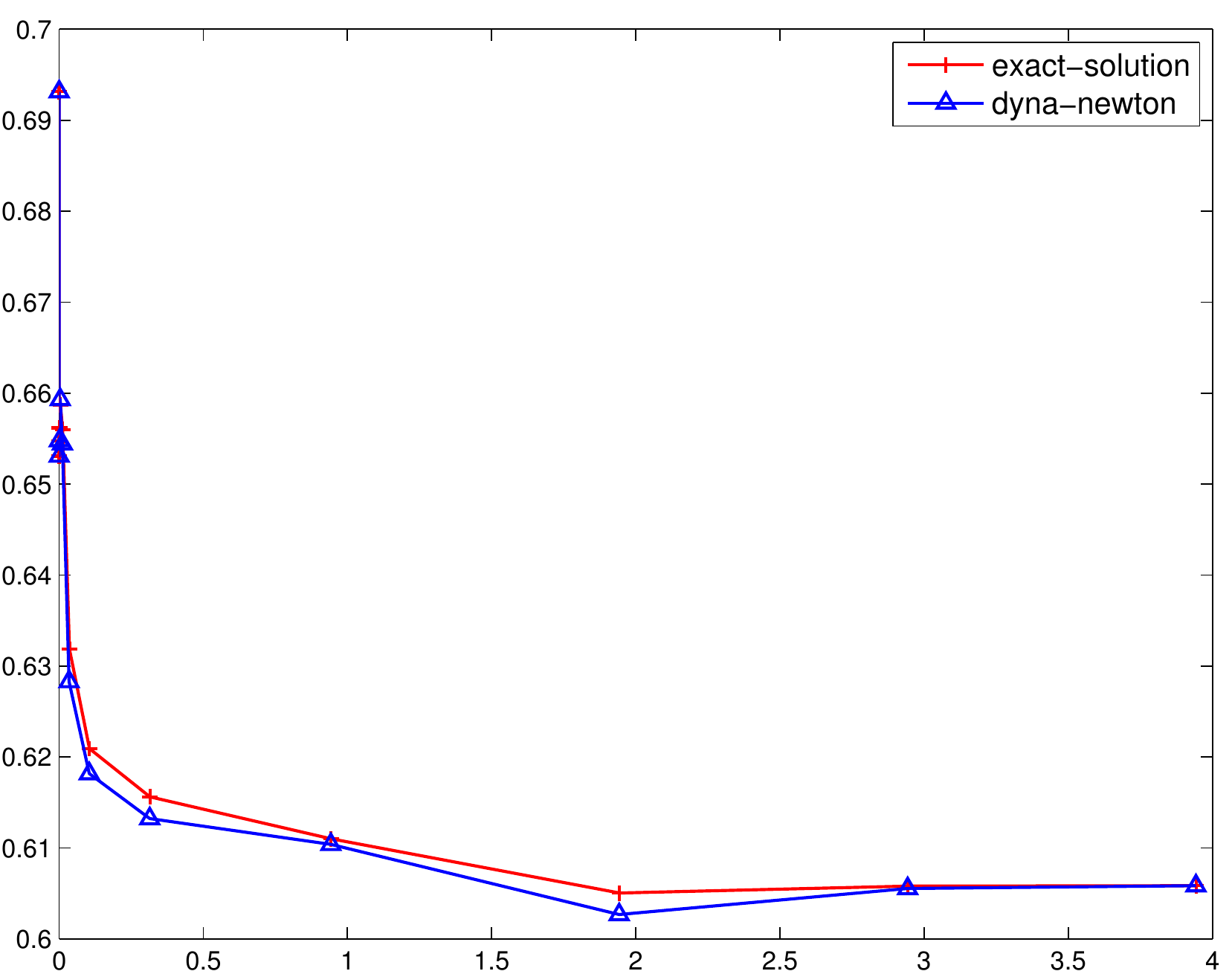}
            \\ 
          3. Test error on {\sc a9a} &
            4. Test error on {\sc covtype} 
	  \end{tabular}
          \caption{{\it Effect of the approximate solution obtained by {\sc dynaNEWTON}}. We here check the effect of the approximate solution obtained by {\sc dynaNEWTON} by iterating for 1 epoch against the exact solution obtained by iterating for 6 epochs.}
          \label{fig:results_approximation_exact}
	\end{center}
\end{figure*}

The analysis we developed assumes that we reach $\x^*_\mu := \arg\min_\x f_{\mu}(\x)$ before switching to $\nu$. We empirically check the robustness to an approximate solution $\hat{\x}_\mu$ obtained by performing a single Newton step instead of $6$ steps, which guarantees convergence up to numerical precision (see Section 9.5 in~\cite{boyd2004convex}). As shown in Figure~\ref{fig:results_approximation_exact}, the impact of the resulting approximate solution is almost negligible.

\subsection{BFGS}

One shortcoming of Newton's method is that it requires solving a linear equation system involving the Hessian matrix, which may be impractical for large and high-dimensional datasets. Approximate variants known as quasi-Newton methods~\cite{dennis1977} have thus been developed, such as the popular BFGS or its limited memory version L-BFGS~\cite{liu1989}. Quasi-Newton methods such as BFGS do not require computing the Hessian matrix but instead construct a quadratic model of the objective function by successive evaluations of the gradient.  
There seems to be a gap between theoretically guaranteed convergence rates and the empirically observed effectiveness of BFGS, in particular on ill-conditioned problems \cite{mokhtari2014b} and for non-convex problems \cite{dean2012large}. 

\begin{figure*}
	\begin{center}
          \begin{tabular}{@{\hspace{5mm}}c@{\hspace{5mm}}c}
            \includegraphics[
            width=0.45\linewidth]{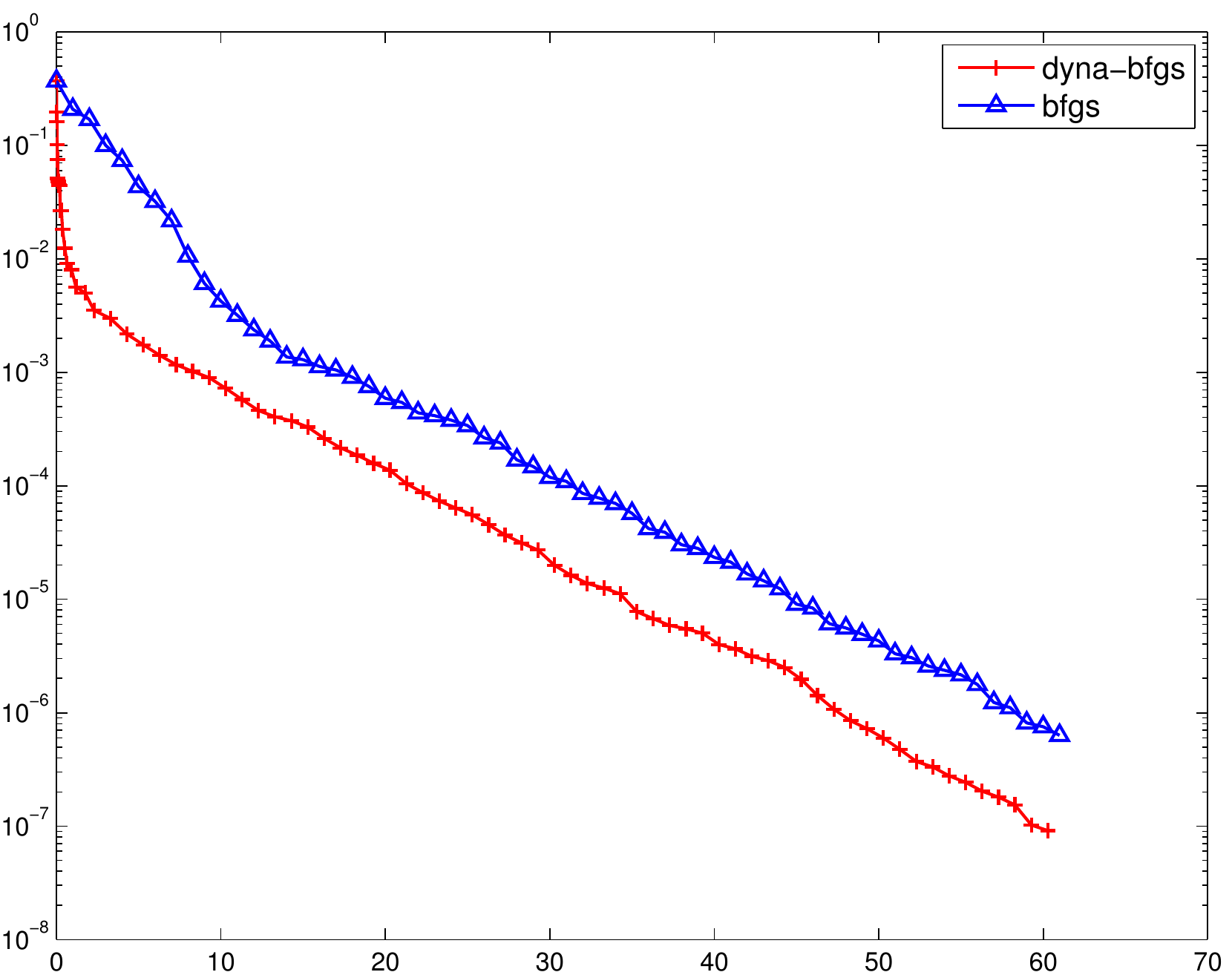} & 
            \includegraphics[
            width=0.45\linewidth]{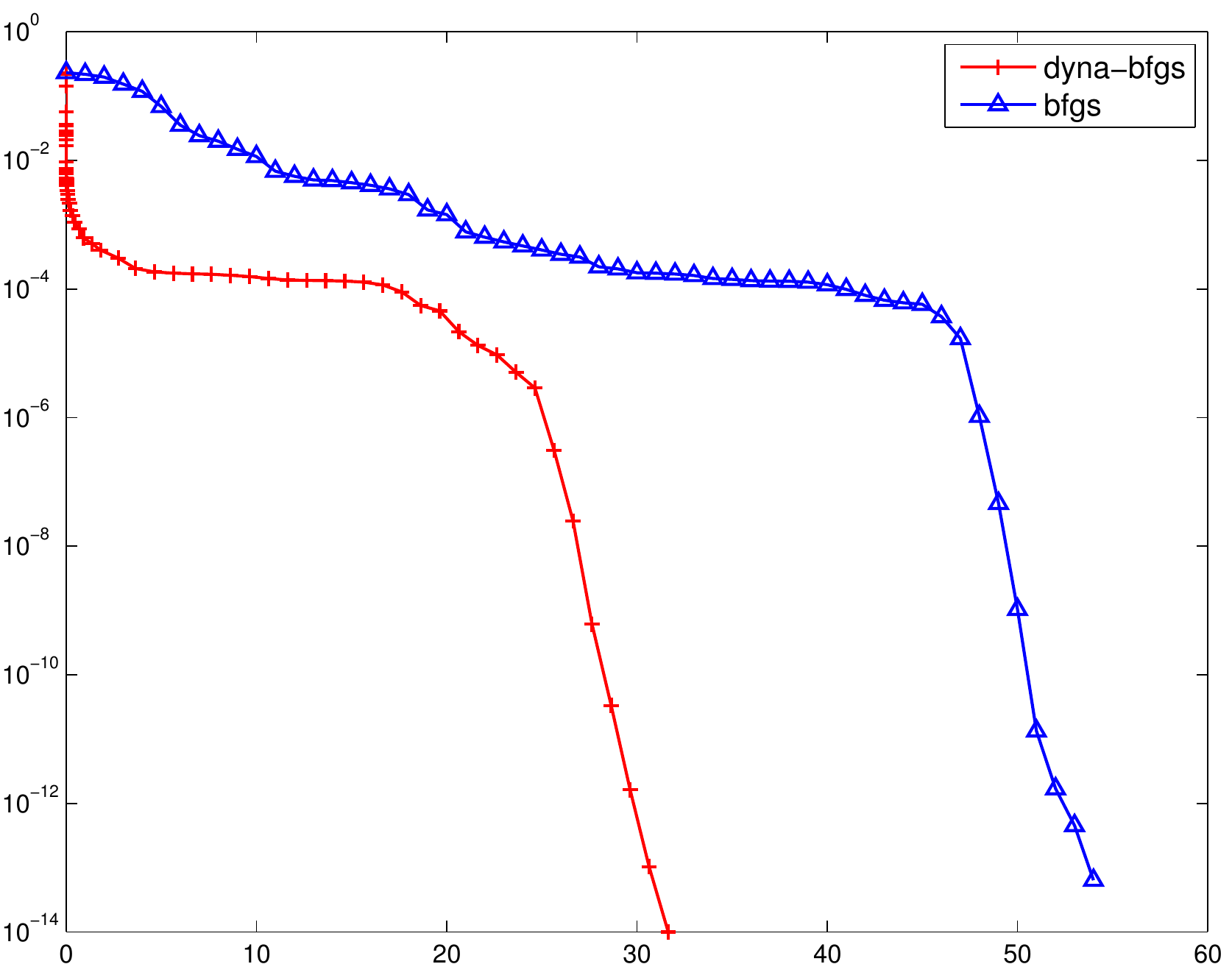}  \\
                1. Empirical suboptimality on {\sc a9a}       &
                 2. Empirical suboptimality on {\sc susy}  
                \\
               \includegraphics[width=0.45\linewidth]{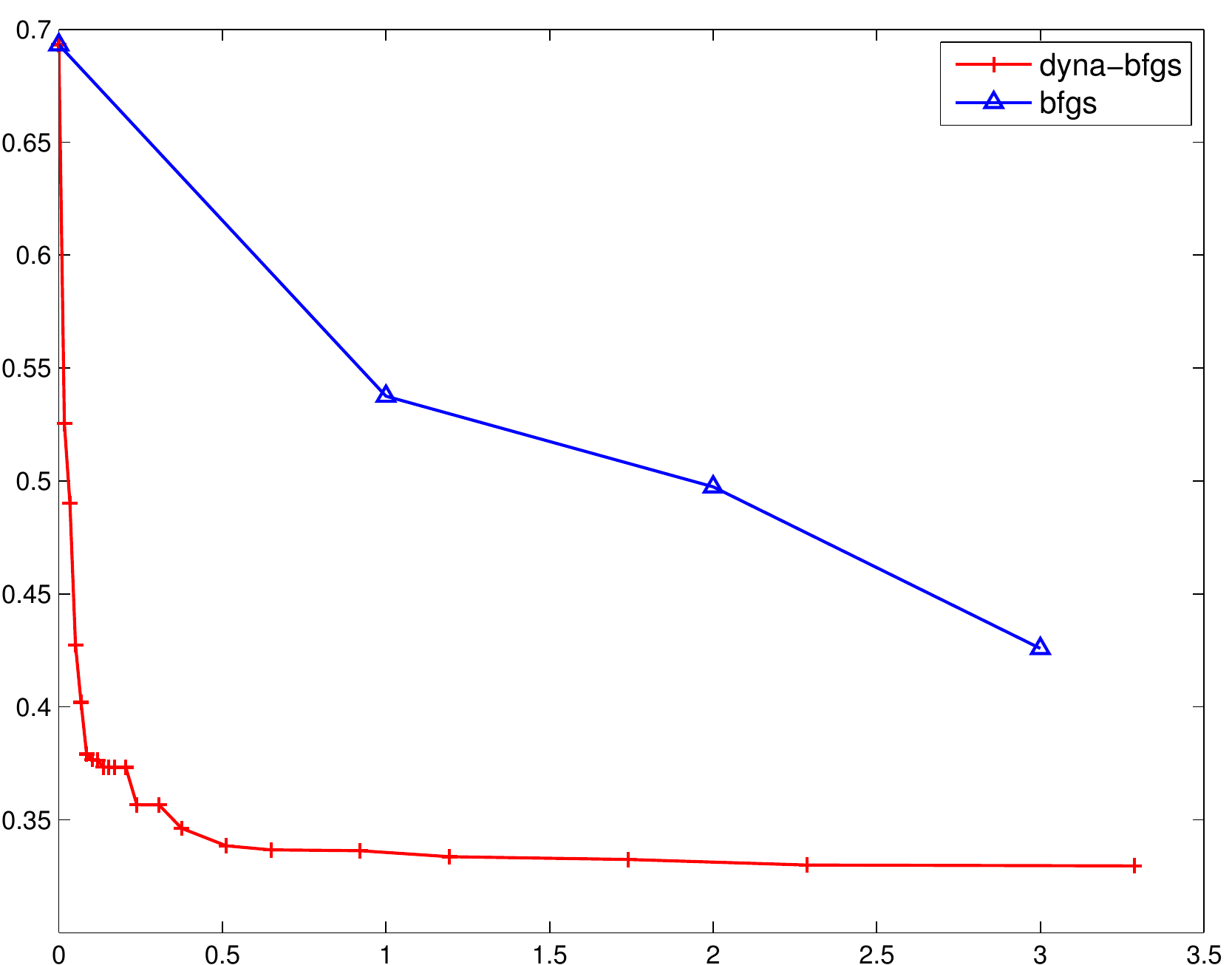} &
            \includegraphics[width=0.45\linewidth]{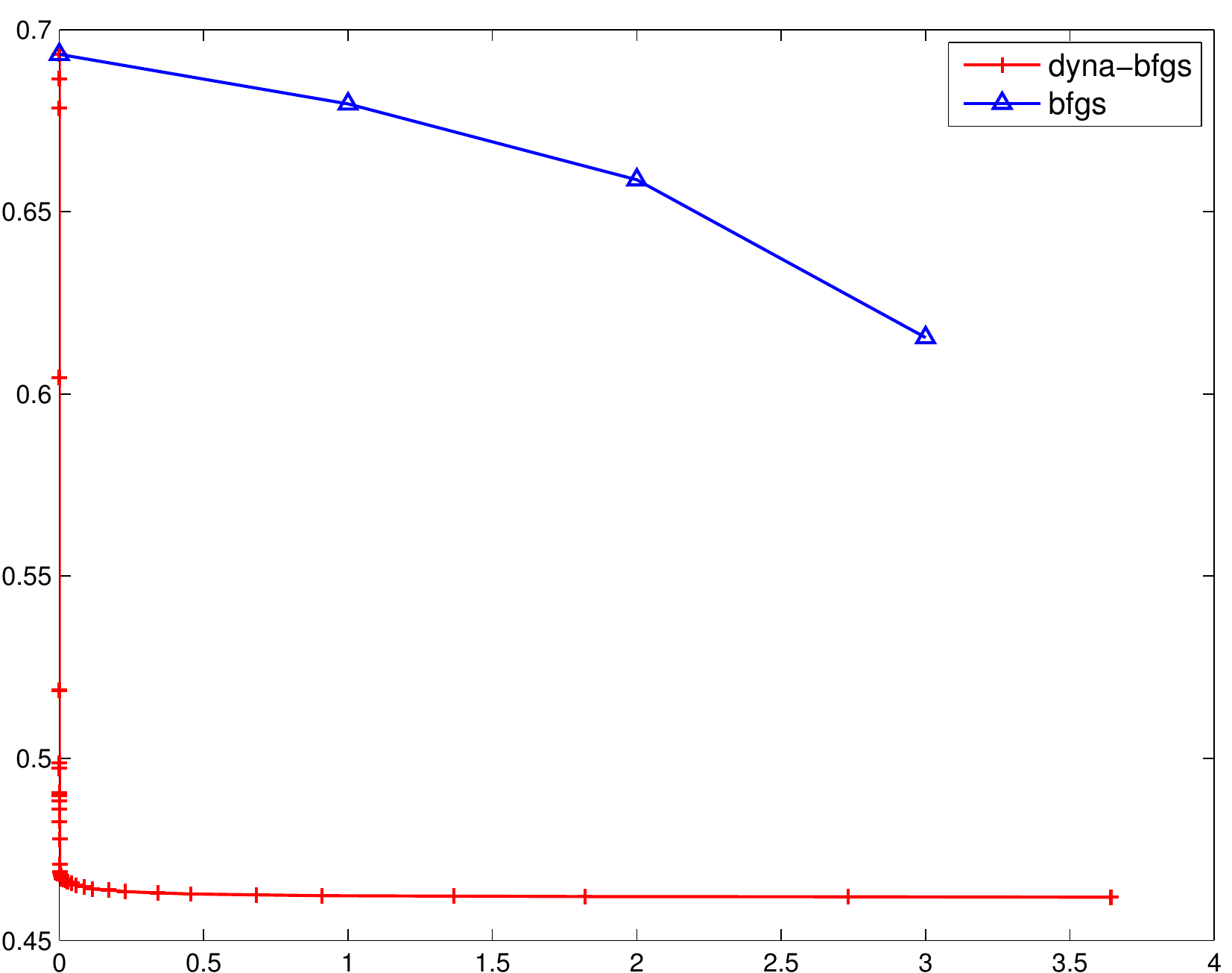}
            \\ 
          3. Test error on {\sc a9a} &
            4. Test error on {\sc susy} 
	  \end{tabular}
          \caption{{\it Comparison of BFGS vs {\sc dynaLBFGS}}. We here provide the empirical suboptimality as well as the test error on the {\sc a9a} and {\sc susy} datasets. We would like to point out that the range of the y-axis between the top and bottom row is different as convergence on the test set was typically observed after 1 or 2 epochs.}
          \label{fig:results_bfgs_epochs}
	\end{center}
\end{figure*}

\begin{figure*}
	\begin{center}
          \begin{tabular}{@{\hspace{5mm}}c@{\hspace{5mm}}c}
            \includegraphics[
            width=0.49\linewidth]{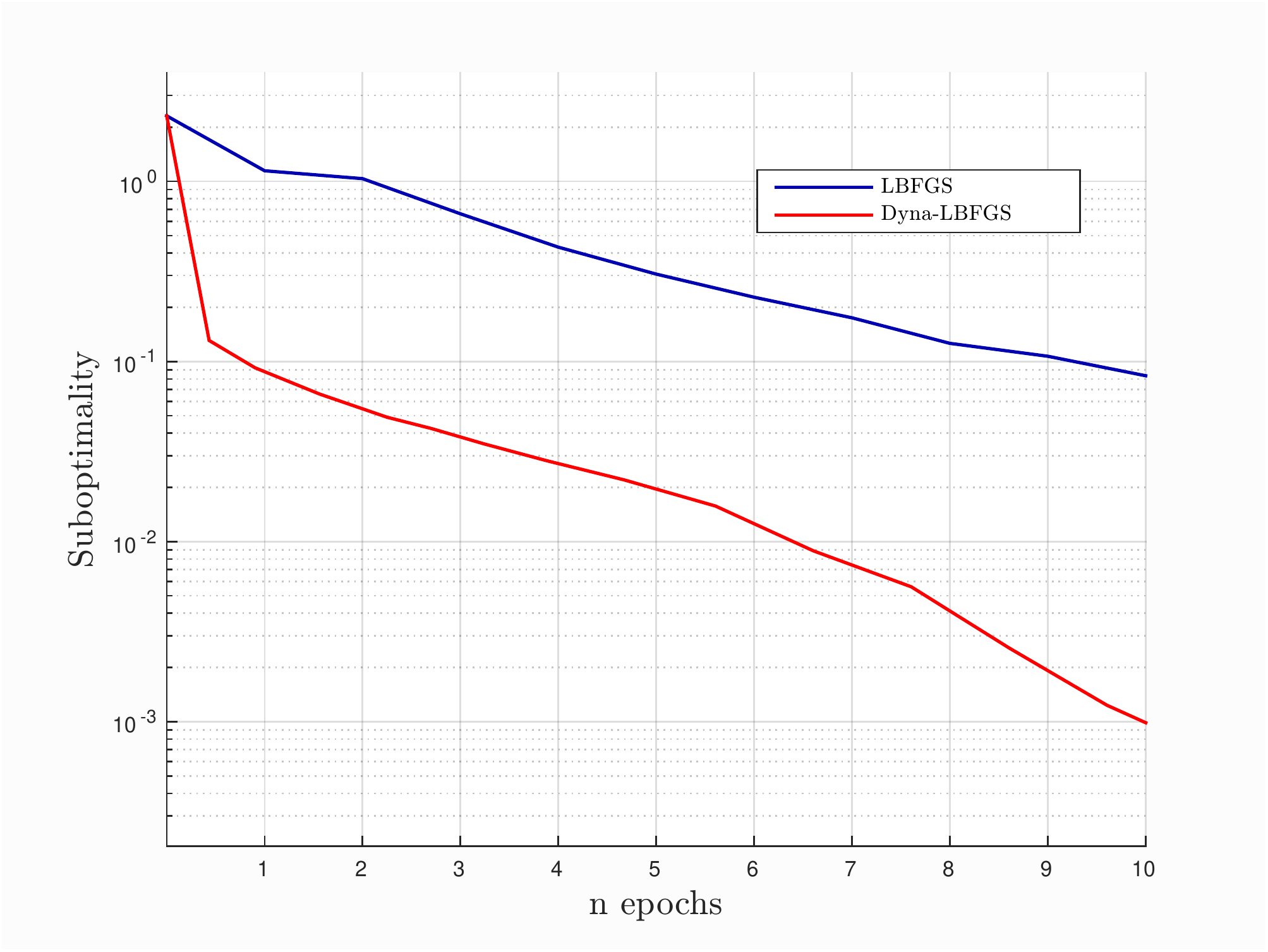} & 
            \includegraphics[
            width=0.49\linewidth]{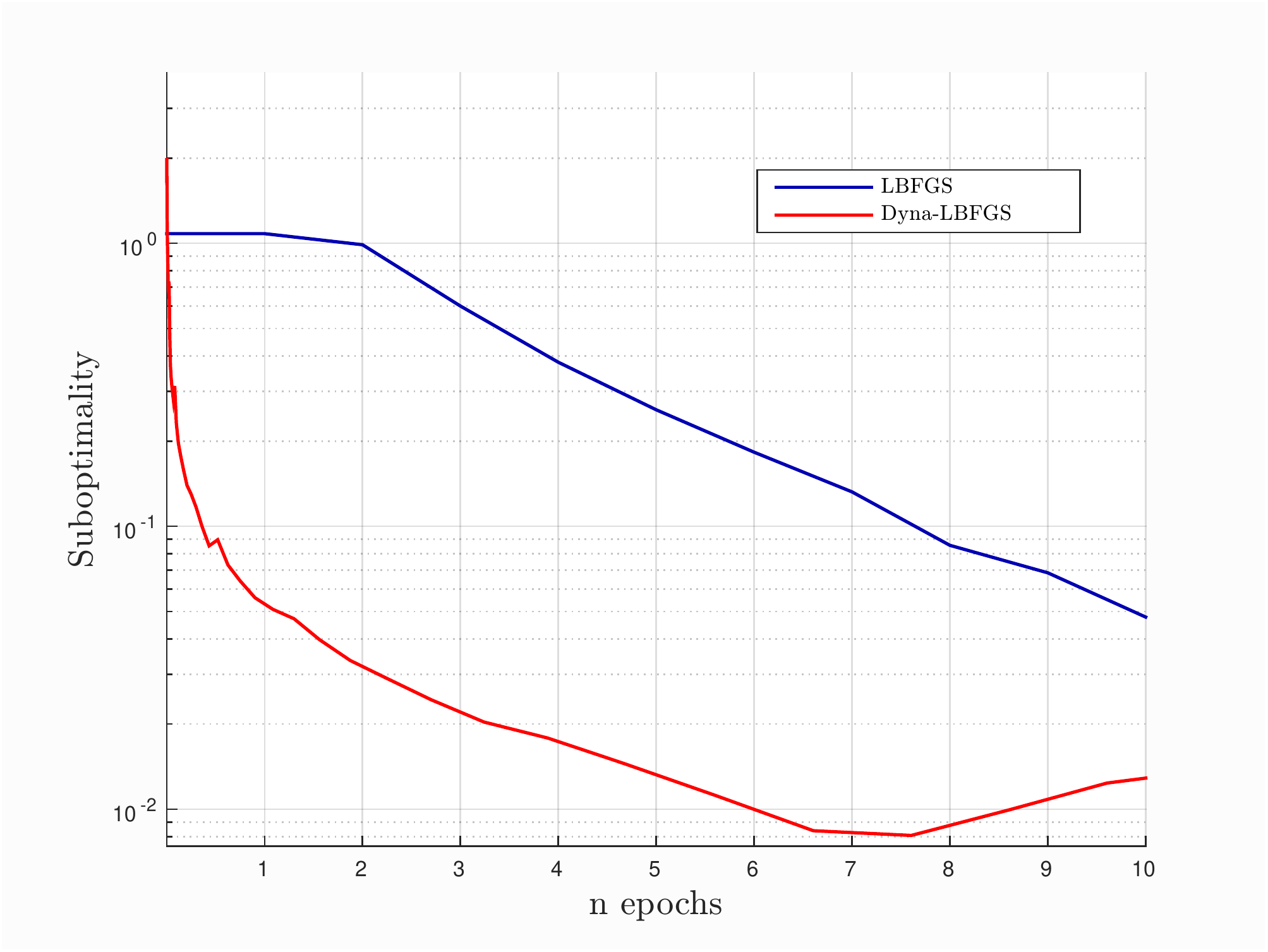}  \\
                1. Training error      &
                2. Test error
	  \end{tabular}
         \caption{{\it Comparison of BFGS vs {\sc dynaBFGS} for training neural networks}. We here show the performance of {\sc dynaBFGS} to train a standard convolutional neural network on the MNIST dataset.}
          \label{fig:results_BFGS_deepnets}
	\end{center}
\end{figure*}

We used the inspiration provided by our continuation method approach to develop a variant of L-BFGS that like {\sc dynaNEWTON}, adaptively changes the sample size and the regularizer. We name this approach {\sc dynaLBFGS}. The pseudo-code of this method is the same as Algorithm~\ref{alg:data_adaptive_newton} except that the Newton increment is computed from the approximate L-BFGS Hessian matrix instead of the true Hessian. We evaluate the performance of {\sc dynaLBFGS} for the task of $\ell_2$-regularized logistic regression on the two datasets described in the main paper. The results shown in Figure~\ref{fig:results_bfgs_epochs} demonstrate significant gains compared to L-BFGS. We also investigate the performance of {\sc dynaLBFGS} on training a convolutional neural network consisting of two convolutional and pooling layers with one fully-connected layer. We include results on the standard MNIST dataset in Figure~\ref{fig:results_BFGS_deepnets}. Although our analysis does not extend to non-convex functions, we nevertheless still observe significant gains.

\end{document}